%% file: ejs-sample.tex
\documentclass[ejsv2]{imsart}

\RequirePackage[numbers]{natbib}
\RequirePackage[colorlinks,citecolor=blue,urlcolor=blue]{hyperref}
\RequirePackage{graphicx}

\usepackage{microtype}
\usepackage{booktabs} 
\usepackage{xcolor}

\usepackage{caption}

\usepackage{mathtools}
\usepackage{algorithm}
\usepackage{algorithmic}
\usepackage{amsfonts}       
\usepackage{bm} 
\input{Notation}

\startlocaldefs
\theoremstyle{plain}

\newtheorem{theorem}{Theorem}[section]
\newtheorem{lemma}[theorem]{Lemma}
\newtheorem{remark}[theorem]{Remark}
\newtheorem{corollary}[theorem]{Corollary}

\theoremstyle{definition}

\theoremstyle{remark}

\endlocaldefs

\begin{document}
\begin{frontmatter}
\title{Regularization can make diffusion models more efficient}
\runtitle{Regularization in diffusion models}

\begin{aug}
\author[A]{\fnms{Mahsa}~\snm{Taheri}\ead[label=e1]{}}
and
\author[A]{\fnms{Johannes}~\snm{Lederer}\ead[label=e2]{}}
\address[A]{Department of Mathematics, University of Hamburg\printead[presep={\\}]{e1,e2}}

\runauthor{M. Taheri \& J. Lederer}
\end{aug}

\begin{abstract}
Diffusion models are one of the key architectures of generative AI. 
Their main drawback, however, is the computational costs.
This study indicates that the concept of sparsity, well known especially in statistics, can provide a pathway to more efficient diffusion pipelines.
Our mathematical guarantees prove that sparsity can reduce the   input dimension's influence on the computational complexity to that of a much smaller intrinsic dimension of the data.
Our empirical findings confirm that inducing sparsity can indeed lead to better samples at a lower cost. 
\end{abstract}



\end{frontmatter}

\section{Introduction}
\input{Contents/body}

\section{Related work}\label{sec:ReWork}
\input{Contents/RelatedWorks}
\section{Empirical support}\label{sec:sim}
\input{Contents/Simulations}

\section{Conclusion}\label{sec:conc}
\input{Contents/conclusion}

\begin{appendix}

\section{Additional simulations and technical results}\label{sec:ComSim}
Here, we present additional simulations and technical results. 

    \input{Contents/Further-Exp}

\input{Contents/Proofs}

\end{appendix}


\begin{funding}
J. Lederer and M. Taheri are grateful for partial funding by the Deutsche Forschungsgemeinschaft (DFG,
German Research Foundation) under project numbers 541176257 and 520388526 (TRR391). 
\end{funding}

\bibliographystyle{imsart-number} 
\bibliography{Contents/References}       


\end{document}

%% file: Notation.tex
\newtheorem{assumption}{Assumption}[section] 

 \newcommand{\draftcolor}{black}

\newcommand{\Dim}{\textcolor{\draftcolor}{\ensuremath{d}}}

\newcommand{\error}{\textcolor{\draftcolor}{\ensuremath{\tau}}}
\newcommand{\R}{\textcolor{\draftcolor}{\ensuremath{\mathbb{R}}}}
\newcommand{\Fx}{\textcolor{\draftcolor}{\x}}
\newcommand{\T}{\textcolor{\draftcolor}{\ensuremath{T}}}
\newcommand{\BY}{\textcolor{\draftcolor}{\widetilde{\x}}}

\newcommand{\q}{\textcolor{\draftcolor}{\ensuremath{q}}}
\newcommand{\dn}{\textcolor{\draftcolor}{\ensuremath{n}}}
\newcommand{\dataset}{\textcolor{\draftcolor}{\ensuremath{\mathcal{D}_n}}}
\newcommand{\x}{\textcolor{\draftcolor}{\ensuremath{\boldsymbol{x}}}}

\newcommand{\score}{\textcolor{\draftcolor}{\ensuremath{\boldsymbol{s}}}}
\newcommand{\scoref}{\textcolor{\draftcolor}{\ensuremath{\score_{\NetP}}}}
\newcommand{\scorefh}{\textcolor{\draftcolor}{\ensuremath{\score_{\DnSMR}}}}

\newcommand{\scorefS}{\textcolor{\draftcolor}{\ensuremath{\score_{\NetPS}}}}
\newcommand{\NetP}{\textcolor{\draftcolor}{\ensuremath{\Theta}}}
\newcommand{\NetPS}{\textcolor{\draftcolor}{\ensuremath{\NetP^*}}}
\newcommand{\E}{\textcolor{\draftcolor}{\ensuremath{\mathbb{E}}}}
\newcommand{\norm}[1]{|\!|#1|\!|}
\newcommand{\normoneM}[1]{|\!|\!|#1|\!|\!|_1}

\newcommand{\varnoise}{\textcolor{\draftcolor}{\ensuremath{\sigma}}}
\newcommand{\tuning}{\textcolor{\draftcolor}{\ensuremath{r}}}
\newcommand{\tuningorc}{\textcolor{\draftcolor}{\ensuremath{r^*}}}

\newcommand{\DnSMR}{\textcolor{\draftcolor}{\ensuremath{\widehat{\NetP}_{\ell_1}}}}

\newcommand{\StandNormal}{\textcolor{\draftcolor}{\ensuremath{\mathcal{N}(\zerov,\mathbb{I}_{\Dim})}}}
\newcommand{\paramspace}{\textcolor{\draftcolor}{\ensuremath{\mathcal{B}}}}
\newcommand{\paramspaceone}{\textcolor{\draftcolor}{\ensuremath{\mathcal{B}_1}}}
\newcommand{\ddpm}{\textcolor{\draftcolor}{DDPMs}}

\newcommand{\sgm}{\textcolor{\draftcolor}{SGMs}}
\newcommand{\sde}{\textcolor{\draftcolor}{Score SDEs}}

\newcommand{\Identity}{\textcolor{\draftcolor}{\ensuremath{\mathbb{I}_{\Dim}}}}

\newcommand{\scaleM}{\ensuremath{\hat{\kappa}}}

\makeatletter
\newcommand{\deq}{\mathrel{\rlap{%
\raisebox{0.3ex}{$\m@th\cdot$}}%
\raisebox{-0.3ex}{$\m@th\cdot$}}=}

\newcommand{\eqd}{=\mathrel{\rlap{%
\raisebox{0.3ex}{$\m@th\cdot$}}%
\raisebox{-0.3ex}{$\m@th\cdot$}}}
\makeatother

\usepackage{tikz}
  \usetikzlibrary{arrows.meta}
  \usetikzlibrary{positioning}
  \tikzset{
    included node/.style={circle, draw=black!100, thick, on grid, minimum width=0.5cm}, 
    hidden node/.style={circle, draw=black!30, thick, on grid, minimum width=0.5cm},
    included connection/.style={->, thick, draw=black!100},
    hidden connection/.style={->, thick, draw=black!30, draw opacity=0},
    fused connection/.style={->, thick, draw=black!100},
    fidden connection/.style={->, thick, draw=black!30, draw opacity=0},
    node title/.style={above=0.8cm of five-one, font=\bfseries},
    general/.style={node distance=1cm and 0.7cm}
}
\newcommand{\xz}{\boldsymbol{x}}
\newcommand{\xt}{\boldsymbol{x}_t}
\newcommand{\Fd}{\mathbb{Q}}

\newcommand{\Bd}{P}
\newcommand{\Fdd}{q}
\newcommand{\Fdds}{\Fdd^s}
\newcommand{\Bdd}{p}
\newcommand{\Bdds}{p^s}
\newcommand{\Bds}{\Bd^s}
\newcommand{\BdA}{\widehat{P}}
\newcommand{\BddA}{\hat{p}}
\newcommand{\KL}{\operatorname{KL}}
\newcommand{\TV}{\operatorname{TV}}
\newcommand{\st}{\boldsymbol{s}_t}

\newcommand{\ut}{\boldsymbol{u}_t}
\newcommand{\uth}{\hat{\boldsymbol{u}}_t}
\newcommand{\us}{\boldsymbol{u}^s}
\newcommand{\scale}{\kappa}
\newcommand{\scaleS}{\kappa^*}
\newcommand{\scaleh}{\hat{\scale}}
\newcommand{\Rs}{\mathcal{R}}

\newcommand{\FSM}{M}

\newcommand{\argmin}{\ensuremath{\arg\,\min}}
\newcommand{\zerov}{\boldsymbol{0}_{\Dim}}
\newcommand{\DerBound}{B}

\newcommand{\batchS}{b_{\operatorname{s}}}
\newcommand{\Esparsity}{\epsilon}
\newcommand{\mnist}{\texttt{MNIST}}
\newcommand{\fmnist}{\texttt{FashionMNIST}}
\newcommand{\cifar}{\texttt{CIFAR10}}
\newcommand{\ButF}{\texttt{Butterflies}}
\newcommand{\deltaT}{\ensuremath{\Delta_{T} (\log \Fdd,\log \Fdds)}}

\newcommand{\ror}{r^*}

%% file: Contents/body.tex
Diffusion models are probabilistic generative models that generate new data similar to those they are trained on~\cite{song2019generative,ho2020denoising,song2020score}.
These models have recently gained significant attention due to their impressive performance    at image generation, video synthesis, text-to-image translation, and molecular design~\cite{dhariwal2021diffusion,ho2022video,ramesh2022hierarchical,xu2022geodiff}.


A diffusion generative model is based on two stochastic processes: 
\begin{enumerate}
    \item A forward process 
$ \Fx_0\to \Fx_1\to \dots \to \Fx_{\T}$ that starts from a sample $\Fx_0 \in \R^{\Dim}$  from a target data distribution~$\q_0$ ($\Fx_0\sim \q_0$) and then diffuses this sample in $\T$ steps into  pure noise $\Fx_{\T}\in  \R^{\Dim}$ ($\Fx_{\T}\sim\StandNormal$).
\item 
A reverse process $\BY_{\T}\to \BY_{\T-1}\to \dots \to \BY_0$ that starts from  pure noise  $\BY_{\T} \in\R^{\Dim}$ ($\BY_{\T}\sim\StandNormal$) and then converts this noise in $\T$ steps into a new sample $\BY_0\in \R^{\Dim}$ that is similar in  distribution to the target distribution.
\end{enumerate}
\noindent 
Making the data noisy is easy.
Therefore, fitting a good reverse process is the key to successful diffusion modeling.

The three predominant formulations of diffusion models are denoising diffusion probabilistic models
(DDPM)~\cite{ho2020denoising}, score-based generative models (SGM)~\cite{song2019generative}, and score-based stochastic differential equations (SDE)~\cite{song2020score,song2021maximum}.
DDPM include two Markov chains: 
a forward process that transforms data into noise, and a reverse process that recovers the data from the noise. 
The objective is to train a function  (usually a deep neural network) for denoising the data over time.
Sample generation then  takes random Gaussian noise through the trained denoising function. 
SGM, which is the setting adopted in  this paper,  also  perturb data with a sequence of  Gaussian noise but then try to estimate the score functions, the gradient of the log probability density, for the noisy data. 
Sampling combines the trained scores with score-based sampling approaches like Langevin dynamics. 
While DDPM and SGM focus on discrete time steps,~\sde~consider infinitely many time steps or unbounded noise levels. In~\sde, the desired score functions are solutions of stochastic differential equations. 
Once the desired score functions are trained,  sampling can be reached using stochastic or ordinary differential equations.

Much research efforts are geared toward non-asymptotic rates of convergence, particularly in the number of steps $\T$ needed to achieve a desired level of reconstruction accuracy.
Typical measures of accuracy are Kullback–Leibler
divergence,  total variation, and Wasserstein distance between the true distribution $\Fd_0$ and the approximated counterpart $\Bd_0$. 
For example, one tries to ensure $\TV(\Fd_0,\Bd_0)\le \error$ for a fixed error level~$\error\in (0,\infty)$, where $\TV(\Fd_0,\Bd_0)\deq \sup_{A\subset \R^d}  |\Fd_0(A)-\Bd_0(A)|$ is called the total variation~\cite{Sara2000}.
The many very recent papers on this topic highlight the large interest in this topic~\cite{block2020generative,de2021diffusion,de2022convergence,lee2022convergence,chen2023probability,chen2023improved,li2023towards,chen2023restoration,liang2024nonN}.
Results like~\cite[Theorem~13]{block2020generative} provide rates of convergence for diffusion models in terms of Wasserstein distance employing Langevin dynamics, 
 but they suffer from the curse of dimensionality in that the rates depend exponentially on the dimensions of the data $\Dim$, that is, the number of input features. 
 Improved convergence rates in terms of $\Dim$ are proposed by \cite{chen2023probability,li2023towards}, showing polynomial growth in~$\Dim$. 
Recently~\cite{liang2024nonN} proposed a new Hessian-based accelerated sampler for the stochastic diffusion processes.
They  achieve  accelerated rate  for the total variation convergence for \ddpm\ of the order $\Dim^{1.5}/\error$ for any target distributions having finite variance and  assuming a uniform bound over the accuracy of the estimated score function (see our Section~\ref{sec:ReWork} for an overview of recent works).
 While these results are a major step forward, they involve a strong dependence on the dimensionality of the data, which is problematic as images, text, and so forth are typically high dimensional. The key question is whether improving these rates with respect to $\Dim$ is possible at all.

 \paragraph{Contribution}  
 This paper aims to enhance the efficiency of diffusion models through the incorporation of regularization techniques commonly used in high-dimensional statistics~\cite{Lederer2021HD}.

The contributions of this work are as follows:  
\begin{itemize}  
    \item We theoretically demonstrate that $\ell_1$-regularization can enhance the convergence rates of diffusion models to  the order of $s^2/\error$, where $s \ll \Dim$, compared to the standard order of $\Dim^2/\error$ (Theorem~\ref{the:maindS}).
    \item We demonstrate that $\ell_1$-regularization can improve sample complexity (Remark~\ref{rem:SC}).
     \item We validate our theoretical findings through  simulations on image datasets (Section~\ref{sec:sim} and Appendix).   
\end{itemize}
\noindent 
Thus, our research is a step forward in the whole field's journey of improving our understanding of  diffusion models and of making diffusion modeling more efficient.

\paragraph{Paper outline}
Section~\ref{sec:disc-setting} introduces score matching and the discrete-time diffusion process.
Section~\ref{Sec:main} presents our proposed estimator along with the main results (Theorem~\ref{the:maindS}).
Section~\ref{sec:ThecRe} includes some technical results and Section~\ref{sec:ReWork} provides an overview of related work.
We support our theoretical findings with numerical observations over image datasets in Section~\ref{sec:sim}.
Finally, we conclude the paper in Section~\ref{sec:conc}.
Additional simulations, technical results, and detailed proofs are provided in the Appendix.

\section{Preliminaries of score matching and discrete-time diffusion process}\label{sec:disc-setting}
In this section, we provide a brief introduction to score matching and  discrete-time diffusion process.
\paragraph{Notations}\label{Notations}
    For a vector $\boldsymbol{z}\in \R^{\Dim}$, we use the notation $\norm{\boldsymbol{z}}_1\deq \sum_{i=1}^{\Dim} |z_i|$, $\norm{\boldsymbol{z}}^2\deq \sum_{i=1}^{\Dim} (z_i)^2$, $\norm{\boldsymbol{z}}_{\infty}\deq \sup_{i\in \{1,\dots,\Dim\}} |z_i|$, and $\norm{\boldsymbol{z}}_0\deq\sum_{i=1}^{\Dim} \mathbf{1}(z_i \neq 0)$. 
   
\subsection{Score matching}
Assume a dataset $\dataset\deq\{\x^1,\dots, \x^{\dn}\}$ of $\dn$ training data  samples $\x^i\in \R^{\Dim}$ with an unknown target distribution $\q_0$ ($\x^i\sim \q_0$ for $i\in\{1,\dots,\dn\}$). 
The goal of probabilistic generative modeling is to use the dataset \dataset\ to learn a model that can sample from~$\q_0$.
The score of a probability density $\q(\x)$, 
the gradient of the log-density with respect to~\x~
 denoted as $\nabla_{\x} \log \q(\x)$,
 are the key components for generating new samples from $\Fdd$.
The score network $\scoref : \R^{\Dim} \to \R^{\Dim}$ is then a neural network parameterized by $\NetP\in \paramspace$, which will be trained to approximate the unknown score $\nabla_{\x} \log \q_0(\x)$. 
The corresponding objective functions for learning scores in \sgm~\cite{song2019generative} is then based on~\cite{hyvarinen2005estimation,hyvarinen2007some}
\begin{equation}\label{scorematch}
 \NetPS\in \arg \min_{\NetP\in \paramspace} \E_{\q_0(\x)}\bigl[\norm{\scoref(\x)-\nabla_{\x}\log \q_0(\x)}^2\bigr]\,,   
\end{equation}
which yields the parameters of a neural network  
$\scorefS(\x)$ that approximates the unknown score function $\nabla_{\x}\log \q_0(\x)$.
Of course, 
the objective function in~\eqref{scorematch} entails (i)~an expectation over $\q_0$ and (ii)~the true score~$\nabla_{\x}\log \q_0(\x)$,
which are both not accessible in practice.
The expectation can readily be approximated by an average over the data samples $\dataset$;
replacing the score needs more care~\cite{vincent2011connection,song2020sliced}.
We come back to this point later  in Section~\ref{sec:distime} by representing  a time dependent form of denoising score matching~\cite{vincent2011connection}.
Once the score function is trained, there are various approaches to generate new samples from the target distribution $\q_0$ employing the approximated score.
These include deterministic and stochastic samplers (see~\cite{li2023towards} for an overview), Langevin dynamics among the most popular one~\cite{song2019generative}.

\subsection{Discrete-time diffusion process}\label{sec:distime}
Let  $\xz\in \R^{\Dim}$ be an initial data sample  and $\xt\in \R^{\Dim}$  for a  discrete time step $t\in\{1,\dots,T\}$  be the latent variable  in the diffusion process. 
Let $\Fd_0$ be the initial data distribution, that is, the distribution belonging to the data's density $\Fdd_0$, and let $\Fd_t$ be the marginal latent distribution in time $t$ in the forward process.
We also use the notation $\Fd_{t,t+1} $ as the  joint distribution over the time $t$ to $t+1$ and $\Fd\deq \Fd_{0,\dots,T} $ as the overall joint distribution over the time $T$. 
In the forward process, white Gaussian noise is gradually added to the data with $\x_t=\sqrt{1-\beta_t} \x_{t-1}+\sqrt{\beta_t}\boldsymbol{w}_t$, where $\boldsymbol{w}_t\sim \mathcal{N}(\zerov,\Identity)$ and  $\beta_t\in (0,1)$ captures the ``amount of noise'' that is injected at time step $t$ and are called the noise schedule. 
This can be written as the conditional distribution 
\begin{equation*}\label{eq:forwardP}
    \Fd_{t|t-1}(\x_t|\x_{t-1})=\mathcal{N}(\x_t;\sqrt{1-\beta_t}\x_{t-1},\beta_t\Identity)\,.
\end{equation*}
An immediate result is that
\begin{equation*}
    \Fd_{t}(\x_t|\x)=\mathcal{N}\bigr(\x_t;\sqrt{\bar\alpha_t}\x,(1-\bar\alpha_t)\Identity\bigl)\,,
\end{equation*}
for $\alpha_t=1-\beta_t$ and $\bar\alpha_t=\prod_{i=1}^{t} \alpha_i$. 
For large enough $T$   we have $\Fd_T\approx \mathcal{N}(\zerov,\Identity)$.
We also denote $\Fdd_t(\x_t|\x)$ as the corresponding density of $\Fd_t(\x_t|\x)$ and that 
$\Fdd_t(\x_t)=\int \Fdd_t(\x_t|\x)\Fdd_{0}(\x)d\x$, in which, $\Fdd_0$ is the unknown target density for $\x$.
We also assume that $\Fd_0$ is absolutely continuous w.r.t. the Lebesgue measure and so the absolute continuity is preserved for all 
$t\in \{1,\dots,T\}$ due to the Gaussian nature of the noise. 
The goal of the reverse process in diffusion models is then  to generate samples (approximately) from the distribution~$\Fd_0$
starting from the Gaussian distribution $\x_T\sim \mathcal{N}(\zerov,\Identity)\eqd\Bd_T$. 
Let's first define 
\begin{equation*}
    \ut(\x_t)\deq\frac{1}{\sqrt{\alpha_t}}\bigl(\x_t+(1-\alpha_t)\nabla_{\x_t}\log \Fdd_t(\x_t)\bigr)\,.
\end{equation*}
At each time step, we then consider the reverse process (for sampling), specifically Langevin dynamics, which can generate samples from a probability density using the true score function, as follows:
\begin{equation*}
    \x_{t-1}=\frac{1}{\sqrt{\alpha_t}}\bigl(\x_t+(1-\alpha_t)\nabla_{\x_t}\!\log \Fdd_t(\x_t)\bigr)+\sqrt{\frac{1-\alpha_t}{\alpha_t}}\boldsymbol{z}_t
    =\ut(\x_t)+\sigma_t
\boldsymbol{z}_t\,,
\end{equation*}
for $\boldsymbol{z}_t\sim \mathcal{N}(\zerov,\Identity)$ and $\sigma_t\deq \sqrt{1-\alpha_t/\alpha_t}$.
Let $\Bd_t$ be the marginal distribution of $\x_t$ in the true reverse process, which is the reverse process by employing the true scores $\nabla_{\x_t}\log \Fdd_t(\x_t)$,  and $\Bdd_t$ be the corresponding density. 
Then, the above statement can be written as 
$\Bd_{t-1|t}=\mathcal{N}(\x_{t-1};\ut(\x_t),\sigma_t^2\Identity)$.
But in practice, one does not have access to the true scores $\nabla_{\x_t} \log \Fdd_t(\x_t)$, 
instead, an estimate of it namely $\scoref(\cdot,t)$, which corresponds to a neural network parameterized with a tuple  $\Theta\in \mathcal{B}$ (tuple of weight matrices; we consider ReLU feedforward neural networks with 
$L$ hidden layers and a total of 
$p
$ parameters),  implying  
\begin{equation*}\label{eq:uhat}
 \uth(\x_t)\deq\frac{1}{\sqrt{\alpha_t}}\bigl(\x_t+(1-\alpha_t)\scoref(\x_t,t)\bigr)\,.
\end{equation*}
Let $\BdA_t$ be the marginal distribution of $\x_t$ in the estimated reverse process implying $\BdA_{t-1|t}=\mathcal{N}(\x_{t-1};\uth(\x_t),\sigma_t^2\Identity)$.
For $\Fd_0$   absolutely continuous, we are then interested in measuring the mismatch between $\Fd_0$ and $\BdA_0$ through the
Kullback–Leibler divergence
\begin{equation*}
 \KL(\Fd_0 ||\Bd_0)\deq  \E_{X\sim \Fd_0} \biggl[\log\frac{\Fdd_0(X)}{\Bdd_0(X)}\biggr] \ge 0\,. 
\end{equation*}

\section{Regularizing denoising score matching}\label{Sec:main}

A promising avenue for accelerating sampling in diffusion models is high-dimensional statistics~\cite{Lederer2021HD}.  
High-dimensional statistics is a branch of statistics that deals with many variables.
A key idea in high-dimensional statistics is the concept of sparsity;
broadly speaking,
it means that among those many variables, only few are relevant to a problem at hand.
There are different sparsity-related approaches in deep learning,
such as dropout~\cite{Hinton2012,Molchanov2017,labach2019survey,gomez2019learning} or explicit regularization~\cite{Alvarez16,Feng17,HEBIRI2025106195}. 
The latter approach adds prior functions (``penalties'', ``regularization'') to the objective functions of the estimators.
These penalties push the estimators toward specific parts of the parameter space that correspond to certain assumptions, for example, sparsity in $\ell_0$-norm and/or $\ell_1$-norm~\cite{Lederer2021HD}. 
The benefits of sparsity are well-documented in regression, deep learning, and beyond~\cite{tibshirani1996regression,Eldar:2012,Hastie2015,Neyshabur2015,Golowich17,Hieber2017,HEBIRI2025106195,taheri2022balancing,mohades2023reducing,golestaneh2024many,taheri2022statistical}.
However, sparsity-inducing prior functions are abundant in statistics and machine learning, 
they are rarely employed for generative models~\cite{lin2016estimation}. 
In this paper, we examine the advantages of incorporating regularization into the objective functions of score-based diffusion models.
Additionally, we leverage techniques from empirical process theory~\cite{Sara2000,Vershynin2018} to analyze regularized objectives and to calibrate the tuning parameter.

\subsection{Denoising score matching under $\ell_1$-regularization}\label{sec:elloneReg}
Here we propose an $\ell_1$-regularized estimator for diffusion models, inspired by the concept of ``scale regularization''  in deep learning~\cite{taheri2021}. Consider the parameter space
\begin{equation}
\label{eq:paramspace}
  \paramspaceone \deq \bigl\{\NetP \in \R^{p}: \normoneM{\Theta}\le 1, ~\norm{\scoref(\x_t,t)}_1~\le~1~~~\forall \x_t\in \R^{\Dim}, t\in \{1,\dots,T\} \bigr\}\,,
\end{equation}
where $\scoref(\cdot,\cdot): (\R^{\Dim},\mathbb{N}) \to \R^{\Dim}$ is modeled as a neural network with two inputs, parameterized by a tuple $\NetP = (W_0, \dots, W_L)$. Here, $\NetP$ collects all the weight matrices of the network, which has $L$ hidden layers and input and output dimensions in $\R^{\Dim}$. 
We consider $\scoref(\cdot,\cdot)$ as a time-dependent score-based model approximating $\nabla_{\x_t} \log \Fdd_t(\x_t)$, which is crucial for sample generation in the backward process of diffusion models. 
The parameter space $\paramspaceone$ corresponds to sparse networks and sparse score functions, meaning that both the network outputs and the parameters are sparse.
Motivated by the denoising score matching objective, a scalable alternative to the objective function in~\eqref{scorematch}~\cite{vincent2011connection}, we define a regularized denoising score-matching estimator as
\begin{equation}\label{dscorematchR}
 (\DnSMR,\scaleM)\in \argmin_{\substack{\NetP\in \paramspaceone \\ \scale\in (0,\infty)}}\E_{\substack{t\sim \mathcal{U}_{[0,T]}\\X_t\sim\Fd_t}}\norm{\scale\scoref(X_t,t)-\nabla_{X_t}\!\log\Fdd_t(X_t)}^2+\tuning \scale^2 \,, 
\end{equation}
where $\scale \in (0,\infty)$ represents the scale of the score function, $\tuning \in [0,\infty)$ is a tuning parameter that balances the penalty between scale and score matching, 
and $\mathcal{U}_{[0,T]}$ denotes the uniform distribution over $[0,T]$.
The fixed constraint $\NetP\in \paramspaceone$ enforces $\ell_1$-norm regularization, while the actual regularization concerns only on the scale $\scale\in (0,\infty)$.
Additionally, we regularize $\scale^2$  to simplify our proofs.
We will further elaborate in Section~\ref{sec:TrainingAlg}  on how the objective function in~\eqref{dscorematchR} can be computed in practice in terms of expectation and score functions.

Our main contribution in this paper is to theoretically and numerically demonstrate that our proposed regularized estimator in~\eqref{dscorematchR} can accelerate the sampling process of diffusion models, specifically increasing the rate of convergence in  Kullback–Leibler divergence from $\Dim^2/\error$~\cite{li2023towards} or $\Dim^{1.5}/\error$~\cite{liang2024nonN} to $s^2/\error$, where $s \ll \Dim$.

We are now ready to introduce some assumptions and present our main theorem for our proposed estimator in~\eqref{dscorematchR}.

We first set the learning rates to be used for our theory and analyses.
For sufficiently large $T$, we set the step size $\alpha_t$ as
\begin{align}\label{eq:stepsize}
    1-\alpha_t\le  c\frac{\log T}{T}\,,~~\forall t\in \{1,\dots,T\}\,,
\end{align}
for a universal constant $c \in(0,\infty)$, which we omit in the remainder of the paper to simplify notation.
We then impose some standard assumptions on the true density function.
    
\begin{assumption}[Finite second moment]\label{Assum:FSM}
There exists a constant $\FSM<\infty$ 
such that $\E_{X_0\sim \Fd_0}\norm{X_0}^2\le \FSM$.    
\end{assumption}
Assumption~\ref{Assum:FSM} simply states that the distribution is not excessively heavy-tailed; it is applied in the proof of our main theorem.
The assumption is standard;
see~\cite{chen2023improved}, \cite{chen2022sampling}, \cite{benton2023linear}, \cite{liang2024nonN}, and many others.

\begin{assumption}[Absolute continuity]\label{assum:conAna}
    We assume that $\Fd_0$ is absolutely continuous w.r.t. the Lebesgue measure, and thus $\Fdd_0$
exists. 
\end{assumption}

We then assume that the derivatives of true log densities are regular, that is, they are bounded by a constant $\DerBound\in (0,\infty)$.
\begin{assumption}[Regular derivatives]\label{assum:ReDe}
For all $t \in \{1,\dots,T\}$ and $l\in \{1,2,\dots\}$ and $\boldsymbol{a}\in [d]^p$ such that $|\boldsymbol{a}|_1=p\in \{1,2,\dots\}$, it holds that
\begin{equation*}
    \E_{X_t\sim \Fd_t}|\partial^p_{\boldsymbol{a}}\log \Fdd_t(X_t)|^\ell ~\le~B~~~\text{and}~~~
    \E_{X_t\sim \Fd_t}|\partial^p_{\boldsymbol{a}}\log \Fdd_t\bigl(\ut(X_t)\bigr)|^\ell~\le~B \,,
\end{equation*}
 for a constant $\DerBound\in (0,\infty)$.    
\end{assumption}
The regularity Assumption~\ref{assum:ReDe} is required for our analysis in Lemma~\ref{lem:RevError} and  also is utilized in previous works like~\cite{huang2024convergence}. As discussed extensively in~\cite[Section~5]{liang2024nonN}, this assumption is relatively mild, for example for distributions with finite variance or   Gaussian mixtures. 
We then push an assumption over the true gradient vectors $\nabla_{\x_t}\log\Fdd_t(\x_t)$ for $t\in \{1,\dots,T\}$, that is, assuming they can be well approximated by some sparse versions.
More precisely, we assume that only a small subset of features or directions in the high-dimensional space contributes significantly to the score functions.
Assumption~\ref{Assum:approxs} is well-motivated in high-dimensional statistics and is central to our main Theorem~\ref{the:maindS}. 
\begin{assumption}[Sparsity]\label{Assum:approxs}
There is a sparsity level $s\in\{1,2,\dots\}$ and an accuracy $\epsilon\in(0,\infty)$, 
such that for all $t \in \{1,\dots,T\}$, there is an analytic auxiliary function $\Fdds_t(\x_t)$ and the corresponding score 
$\nabla_{\x_t}\log\Fdd_t^s(\x_t)$ that is $s$-sparse and $\epsilon$-accurate:
\begin{multline*}
\E_{X_t \sim \Fd_t}\norm{\nabla_{X_t}\log\Fdd_t^s(X_t)}_0~\leq~s~~\text{and} \\
\frac{1}{T} \sum_{t=1}^T
   \sqrt{\E_{X_t \sim \Fd_t} \norm{\nabla_{\x_t}\log\Fdd_t(X_t) - \nabla_{\x_t}\!\log\Fdds_t(X_t)}^2}~\leq~\Esparsity\,.
\end{multline*}
\end{assumption}

Our sparsity assumption above is formulated as an \emph{average} over all time steps $t$.
It states that, among the many features of the space, only a small subset has a major impact on the score vectors on average, while the majority contribute only marginally ($s< \Dim$).
Naturally, the sparsity level $s$ may vary depending on the dataset and could grow at different rates in different contexts. 
We refer the reader to the detailed discussion following Theorem~\ref{the:maindS} and in Section~\ref{sec:assumdisc}, where we analyze the worst-case scenario for the scaling behavior of $\epsilon$ relative to the rate of the tuning parameter $\tuning$, some  simple examples that sparsity holds in practice, and empirically validation of sparsity on MNIST and CIFAR10 data. 
Also, we refer to Assumption~\ref{ass:Rsparsity} in the supplementary material for a more relaxed version of our sparsity assumption, which can be used in place of Assumption~\ref{Assum:approxs}.


\begin{theorem}[Non-asymptotic rates of convergence for regularized diffusion models]\label{the:maindS}
Under the Assumptions~\ref{Assum:FSM},~\ref{assum:conAna},~\ref{assum:ReDe}, and ~\ref{Assum:approxs}  and
    for $\tuning\ge \tuningorc\deq C_{\x}\sqrt{\log (np)/n}$, our (in-sample) estimator proposed in~\eqref{dscorematchR}  generates samples with
    \begin{align*}
     \KL(\Fd_0 ||\BdA_0)
     &~\le~\frac{\FSM}{T^2}+\frac{1}{T}\max\{1,9(s\DerBound)^2\} + C_{\x}s^2\DerBound^2\sqrt{\frac{\log (nTp)}{n}}
      +\Delta_{T} (\log \Fdd,\log \Fdds)\\
     &~~~~~~+\inf_{\substack{\NetP\in \paramspaceone\\ \scale\in (0,\infty)}}\biggl\{\frac{\log T}{T}  \sum_{t=1}^{T}\frac{1}{n} \sum_{i=1}^{n}  \norm{\scale\scoref(\x_t^i,t)-\nabla_{\x_t}\!\log\Fdd_t(\x_t^i)}^2
      +\tuning\scale^2\biggr\}\,,
\end{align*}

where  $\Delta_{T} (\log \Fdd,\log \Fdds)
    \deq \sum_{t=1}^{T} (\E_{X_{t}\sim \Fd_{t}}[\E_{X_{t-1}\sim \Fd_{t-1|t}}[\log \Fdd_{t-1}(X_{t-1})-\log \Fdds_{t-1}(X_{t-1})]\notag-\allowbreak\E_{X_{t-1}\sim \Bds_{t-1|t}}\allowbreak[\log \Fdd_{t-1}(X_{t-1})-\log \Fdds_{t-1}(X_{t-1})]])$, $C_{\x}$  a constant depending on the input distribution, and 
$
p$ denotes the total number of network parameters with probability at least $1-1/n$.
\end{theorem}
\begin{corollary}
[Parametric setting]\label{cor:maindS}
Assume that $\tuning=\tuningorc$ and that there exists a pair $(\Theta^*,\kappa^*)\in \paramspaceone \times (0,\infty)$ such that $\kappa^*\scorefS(\x_t^i,t)=\nabla_{\x_t}\log q_t(\x_t^i)$ for all $i\in\{1,\dots,n\}$ and $t\in \{1,\dots,T\}$. Then, under the assumptions of Theorem~\ref{the:maindS}, our (in sample) estimator proposed in~\eqref{dscorematchR}  generates samples with
    \begin{equation*}
     \KL(\Fd_0 ||\BdA_0)
     ~\le~\frac{1}{T}\max\{1,9(s\DerBound)^2\}+ C_{\x}\sqrt{\frac{\log (nTp)}{n}}\max\{(s\DerBound)^2,(\scale^*)^2\}
      +\Delta_{T} (\log \Fdd,\log \Fdds)
\end{equation*}
 with probability at least $1-1/n$.
\end{corollary}
Our Theorem~\ref{the:maindS} reveals that if the true gradient vectors $\nabla_{\x_t} \log q_t(\x_t)$ can be well approximated by $s$-sparse vectors $\nabla_{\x_t}\log q^s_t(\x_t)$ and for sufficiently large tuning parameter, the rates of convergence of diffusion models scale with $s$, where potentially  $s \ll \Dim$.  
The term $\Delta_{T}(\log \Fdd,\log \Fdds)$ in the bound of Theorem~\ref{the:maindS} measures how close the log density  $\log \Fdd_t(\x_t)$ is to the auxiliary log density $\log \Fdds_t(\x_t)$ across the entire sample space and time steps.  
Note that we use the notation $\Bds_t$ for the distribution of the latent steps in the reverse process, utilizing the sparse scores of Assumption~\ref{Assum:approxs} (see also Section~\ref{sec:Aux} for more details). 
Following the intuition behind score matching, we argue that Assumption~\ref{Assum:approxs} also promotes closeness between these log densities.
While one might argue that the sparsity assumption for the score functions at large time steps $t$ and  in~\ref{Assum:approxs} may not always hold, our detailed discussion in Section~\ref{sec:assumdisc} demonstrates that, due to the carefully chosen order of the tuning parameter, our estimator performs comparably to standard score matching even in the worst-case scenario—namely, when the score functions exhibit no sparsity on average. However, when some degree of sparsity is present, our method can lead to significant improvements by promoting sparse representations while still keeping the score estimation error small.
Put differently, our estimator strikes a balance between the average score estimation error and the level of sparsity. As a result, it not only achieves low estimation error but also identifies and leverages sparsity level (or scale) when it exists.
Furthermore, our empirical observations in Sections~\ref{sec:sim} and~\ref{sec:assumdisc} support the practical validity of our assumptions.
Following the work of~\cite{karras2024analyzing}, we also conjecture that the $\ell_1$-regularizer may serve as a useful tool for improving the training dynamics of diffusion models, although a thorough investigation is still needed.
Theorem~\ref{the:maindS}  also directly implies a bound on the the total-variation distance between $\Fd_0$ and $\BdA_0$ in view of Pinsker's inequality.
Detailed proof of Theorem~\ref{the:maindS} is provided in Appendix~\ref{proof:mainR}. 
Corollary~\ref{cor:maindS} follows directly from Theorem~\ref{the:maindS}, so we omit its proof.

\cite{li2023towards,liang2024nonN}
show that the reverse diffusion process produces a sample with error roughly at the order of $\varsigma$ if  $\sum_{t=1}^T\E_{X_t\sim \Fd_t}\norm{\score(X_t,t)-\nabla_{X_t}\log \Fdd_t(X_t)}^2/T\le \varsigma$. But whether such accurate estimators are available in practice remains unclear.
For example,
\cite[Theorem 3.5, Corollary 3.7]{zhang2024minimax} upper bounds the estimation error of the score (using $n$ training samples) at the order of $n^{-2\beta/(2\beta+\Dim)}$ assuming the true data distribution $\Fdd_0$ is 1.~$\sigma_0$-sub-Gaussian and 2.~in the Sobolev class of density functions with the order of smoothness $\beta \le 2$, see also \cite{wibisono2024optimal,dou2024optimal}. 
This result highlights that the original score matching method suffers from the curse of dimensionality (note that $\beta \le 2\ll \Dim$). 
Thus, regularization not only  accelerate the reverse process but also help in the estimation of the scores directly---compare to~\cite{lederer2023extremes}. 
\cite{block2020generative} and~\cite{gupta2024improved} studied the sample complexity of score matching, providing bounds that scale as $O(d^{5/2}(B'^2p)^L\sqrt{L}/\tau^2)$ and $O(d^{2}(Lp)\log B'/\tau^3)$, respectively. Here, $d$ denotes the data dimensionality, $p$ the total number of network parameters (each bounded by $B'$), and $L$ the number of hidden layers of the network.  In contrast, our result in Corollary~\ref{cor:maindS} (see also Remark~\ref{rem:SC}) shows a sample complexity scaling as $O(\max\{(sB)^4,(\scale^*)^4\}\log (pn)/\tau^2)$. This demonstrates a substantial improvement for our regularized estimator: the bounds now depend on the effective sparsity $s \ll d$ and grow only logarithmically with the total number of network parameters, rather than polynomially.
More precisely, the bound can even decrease exponentially with the number of hidden layers of the network, which we omit here for simplicity; see~\cite{taheri2021}.
In fact, regularization can reduce the effective complexity of the network space under consideration, thereby improving the sample complexity of score estimation. This is consistent with the results of~\cite{zhu2023sample}, who establish convergence rates of order $O(d\log \mathcal{N}(\cdot,\mathcal{F})/\tau^2)$, where $\mathcal{N}(\cdot,\mathcal{F})$ quantifies the complexity of the network function class. 

 \begin{remark}[Sample complexity]\label{rem:SC}
Corollary~\ref{cor:maindS} states that once the network space is sufficiently large (that is, there exists a pair $(\Theta^*, \kappa^*) \in \paramspaceone \times (0,\infty)$ such that $\kappa^* \scorefS(\x_t^i, t) = \nabla_{\x_t} \log q_t(\x_t^i)$ for all $i \in \{1,\dots,n\}$ and $t \in \{1,\dots,T\}$), the sample complexity of the regularized diffusion model increases by 
\[
O\!\left(C_{\x}^2 \frac{\log (nTp)}{\tau^2}\max\big\{(s \DerBound)^4, (\scale^*)^4 \big\}\right)
\]
in order to achieve 
\[
\KL(\Fd_0 \,\Vert\, \BdA_0) \le \tau.
\]
\end{remark}

We highlight that our regularization technique is motivated by two main considerations: 
First, it allows for a more focused search during sample generation by concentrating on the features 
that have the greatest influence on the generated samples (see Figure~\ref{fig:toy}). 
Second, as noted in~\cite[Example~4.1]{ren2025unified}, one of the dominant sources of error in the convergence of diffusion models 
is the estimation error
\begin{equation*}
 \mathcal{E}_{\mathrm{est}}(\Theta)
 = \mathbb{E}_{\x_0\sim q_0}\!\left[\int_0^T 
    \mathbb{E}_{\x_t\sim q_{t|0}}
      \bigl[\|s_{\Theta}(\x_t,t)-\nabla\log q_{t|0}(\x_t|x_0)\|^2\bigr]
    \, dt\right],
\end{equation*}
which, in the finite-sample regime, can lead to overfitting. 
Thus, regularization—being a classical and effective remedy for overfitting—naturally plays a crucial role to reduce estimation error.

\section{Technical results}\label{sec:ThecRe}
Here we provide some auxiliary results  used in the proof of Theorem~\ref{the:maindS}. 

\begin{lemma}[Reverse-step error]\label{lem:RevError}
Under the Assumptions~\ref{assum:ReDe} and~\ref{Assum:approxs}  we have 
 \begin{align*}
    \sum_{t=1}^{T}~\E_{X_t,X_{t-1}\sim \Fd_{t,t-1}}\biggl[\log\frac{\Fdd_{t-1|t}(X_{t-1}|X_t)}{\Bdd^s_{t-1|t}(X_{t-1}|X_t)}\biggr]\notag&\le \frac{1}{T}\bigl(s^2\DerBound^2+  s^2\DerBound^2+ s^2\DerBound^2\epsilon+ s\DerBound \epsilon\bigr)\notag\\
    &~~~~~~+\Delta_{T} (\log \Fdd,\log \Fdds)\,.
\end{align*}
 
\end{lemma}
Lemma~\ref{lem:RevError} helps upper bounding the reverse-step error for the backward process of diffusion models and 
its detailed proof is provided in Appendix~\ref{proof:lemReverror}. 

We then present a lemma that aids in determining the optimal rates for the tuning parameter.
\begin{lemma}[Empirical processes]\label{lem:emp}
Under the Assumption~\ref{Assum:approxs} 
we obtain 
\begin{align*}
   \sup_{\NetP\in \paramspaceone}\Bigl|\E_{X_t\sim Q_t}\norm{\scaleh\scoref(X_t,t)&-\nabla\log q^s_t(X_t)}^2-\frac{1}{n} \sum_{i=1}^{n}  \norm{\scaleh\scoref(\x_t^i,t)-\nabla\log q^s_t(\x_t^i)}^2\Bigr|\\
     &\le C_{\x} (\scaleh^2+s^2 \DerBound^2) \sqrt{\frac{\log (np)}{n}}
\end{align*}
with probability at least $1-32/n$, where $C_{\x}$ is a constant depending on the input distribution, and 
$
p$ denotes the total number of network parameters.
\end{lemma}
Lemma~\ref{lem:emp} is employed in the proof of Theorem~\ref{the:maindS} to calibrate the tuning parameter. 
Its detailed proof  is provided in Appendix~\ref{proof:lememp}.

%% file: Contents/RelatedWorks.tex
The non-asymptotic rates of convergence for diffusion models established very recently~\cite{block2020generative,de2021diffusion,de2022convergence,lee2022convergence,chen2023probability,chen2023improved,li2023towards,chen2023restoration,huang2024convergence,Gitte2025} show the large interest in this topic and are an important step forward but do not fully explain the success of generative models either.
For example, results like~\cite[Theorem~13]{block2020generative} provide rates of convergence for diffusion models in terms of Wasserstein distance employing the tools from empirical-process theory, 
but they suffer from the curse of dimensionality in that the rates depend exponentially on the dimensions of the data, that is, the number of input features. 
 Recent works then concentrate on improving convergence guarantees to grow polynomially in the number of input features under different assumptions on the  original and estimated scores ($L_2$-accurate score estimates, Lipschitz or smooth scores, scores with bounded moments)~\cite{lee2022convergence,wibisono2022convergence,chen2022sampling,chen2023probability,chen2023improved,chen2023restoration,lee2023convergence,huang2024convergence}. 
 For example,~\cite{lee2022convergence} prove a convergence guarantee  in terms of total variation for~\sgm, 
 which has a polynomial dependence on the number of input features if the score estimate is $L_2$-accurate for any smooth distribution satisfying the log-Sobelev inequality.
A very recent work by~\cite{li2023towards} proposes improved convergence rates in terms of total variation for~\ddpm~with ordinary differential equations and stochastic differential equations samplers that are proportional to $\Dim^2/\error$,
where $\Dim$ is the number of input features and $\error$ the error in the measure under consideration.
They assumed 1.~finite support  assumption, 2.~$L_2$-accurate score estimates, and 3.~accurate Jacobian matrices. 
\cite[Theorem~3]{li2023towards} also provides rates growing by~$\Dim^3/\sqrt{\error}$ for an accelerated ordinary differential equations samplers. 

While works like~\cite{li2023towards} and \cite{li2024accelerating} concentrate more on improving the rates in~$\error$,~\cite{chen2023probability} focus on improving the rates in~$\Dim$ for denoising diffusion implicit models.
\cite{chen2023probability} use a specially chosen corrector step based on the underdamped Langevin diffusion to achieve their improvements, namely rates  proportional to $L^2\sqrt{\Dim}/\error$
by assuming: 1.~the score function along the forward process is $L$-Lipschitz,  2.~finite second moments of the data distribution, and  3.~$L_2$-accurate score estimates. 
\cite{chen2023improved,benton2023linear} then  relaxed the assumptions over the data distribution and  proposed  the  rates of convergence for DDPM  proportional to $\sqrt{\Dim^2/\tau}$ and $\sqrt{\Dim/\tau}$ under 1.~finite second moments of the data distribution, and  2.~$L_2$-accurate score estimates. 
Further research directions may also build upon~\cite{chen2023score}, who assume that the data lie on a low-dimensional linear subspace. They demonstrate that, in this setting, the convergence rates depend on the dimension of the subspace.


%% file: Contents/Simulations.tex
In this section, 
we demonstrate the  benefits of regularization for diffusion empirically.
We present a toy example and the MNIST family dataset here, deferring additional simulations and setups to Appendix Section~\ref{sec:ComSim}, where we study more complicated datasets including FashionMNIST, Butterflies, and CIFAR10.

\subsection{Toy example}
\begin{figure*}[htbp]
    \begin{minipage}{0.27\textwidth}
        \centering
        \includegraphics[width=\textwidth]{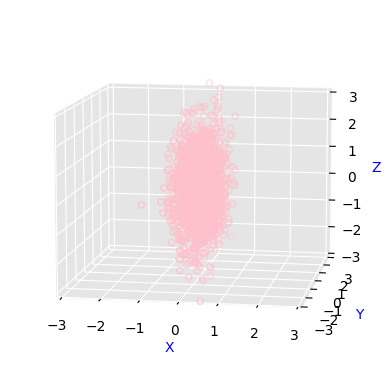}
    \end{minipage}
       \hfill
    \begin{minipage}{0.3\textwidth}
        \centering
        \includegraphics[width=\textwidth]{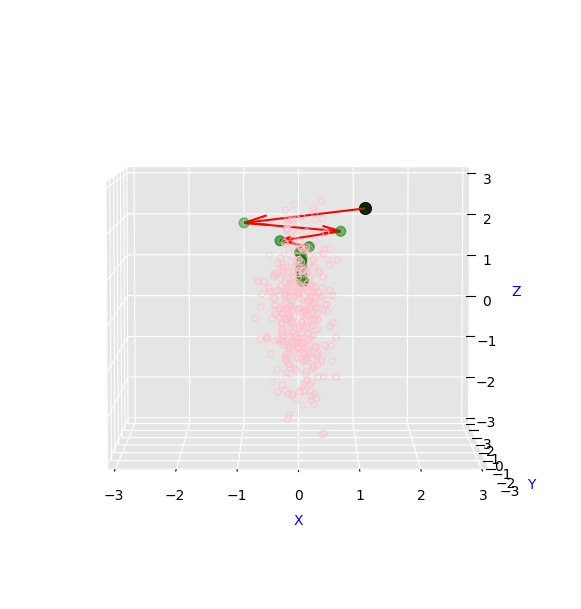}
    \end{minipage}
    \hfill
    \begin{minipage}{0.3\textwidth}
        \centering
        \includegraphics[width=\textwidth]{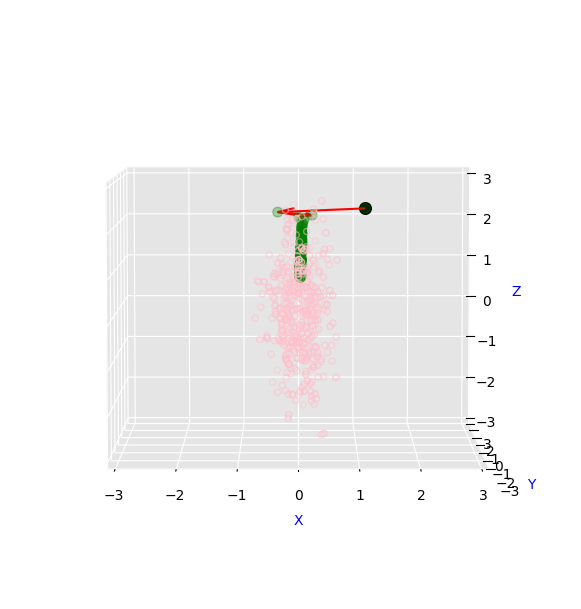}
    \end{minipage}

    \caption{Visualizing the sampling process for 3D data (first panel) with original denoising score matching (second panel) versus regularized denoising score matching (third panel).
The original samples are depicted as red circles, blue circles indicate the starting points for sampling, and green circles represent the latent generated samples.
The red arrows illustrate the sampling paths. It is evident that regularized denoising score matching predominantly adheres to the two-dimensional sub-manifold (along the $Y$ and $Z$ axes), whereas the original denoising score matching explores the entire 3D space.
    }\label{fig:toy}
\end{figure*}

We first highlight the influence of regularization on the sampling process of a 3D toy example. 
We consider $2000$ three-dimensional, independent Gaussian samples with mean zero and covariance matrix $\allowbreak[0.08,0,0;0,1,0;0,0,1]$;
hence, the data fluctuate most around the $y$ and $z$ axes.
We then train two diffusion models, with the same data: 
the original denoising score matching and the same with an additional sparsity-inducing regularization (as proposed in~\eqref{dscorematchR}) and $r=0.001$.
Figure~\ref{fig:toy} visualizes the data (first panel) and the sampling process with $T=60$ for original score matching (second panel) and the regularized version (third panel).
Both models start from the blue dot. 
The figure shows that the regularized version provides a more focused sampling. 
\subsection{MNIST}\label{sec:mnist}
\begin{figure*}[t]
    \centering
    \begin{minipage}{0.15\textwidth}
        \centering
        \includegraphics[width=\textwidth]{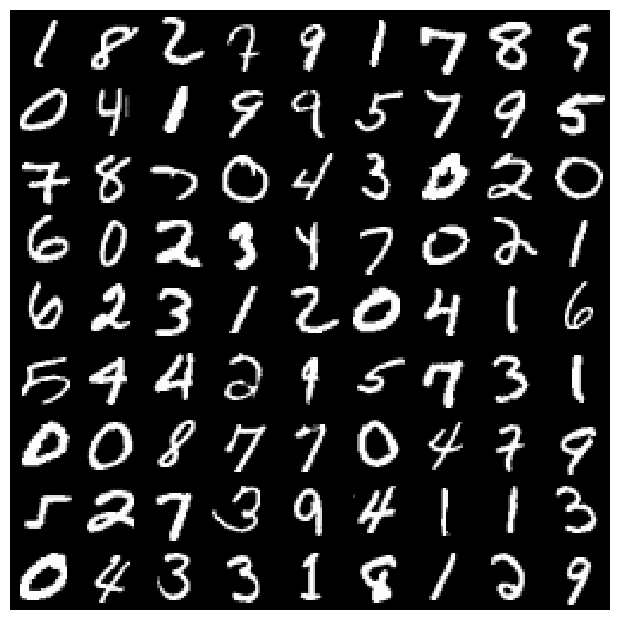}
    \end{minipage}
    
    \vspace{0.2em} 
    
    \begin{minipage}{0.15\textwidth}
        \centering
        \includegraphics[width=\textwidth]{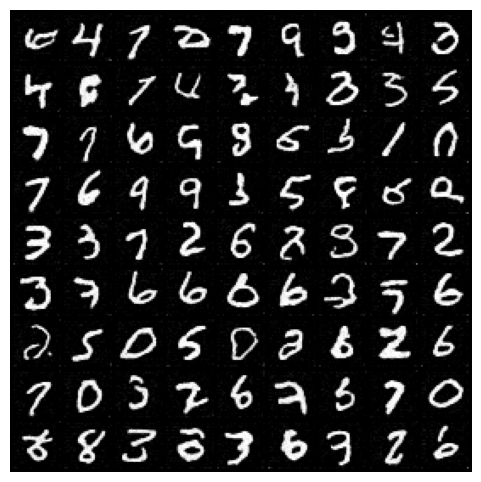}
    \end{minipage}
    \hspace{9em} 
    \begin{minipage}{0.15\textwidth}
        \centering
        \includegraphics[width=\textwidth]{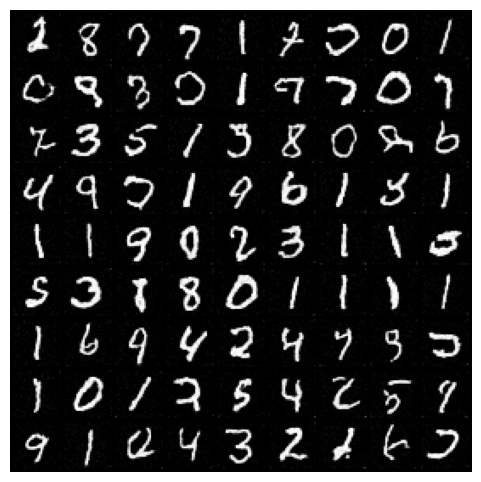}
    \end{minipage}
    
    \vspace{0.2em} 
    \begin{minipage}{0.15\textwidth}
        \centering
        \includegraphics[width=\textwidth]{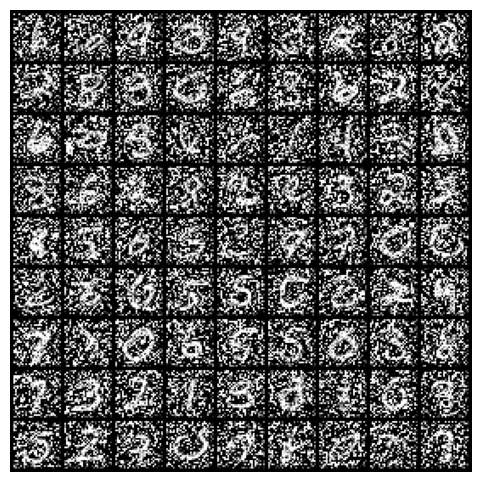}
    \end{minipage}
    \hspace{9em} 
    \begin{minipage}{0.15\textwidth}
        \centering
        \includegraphics[width=\textwidth]{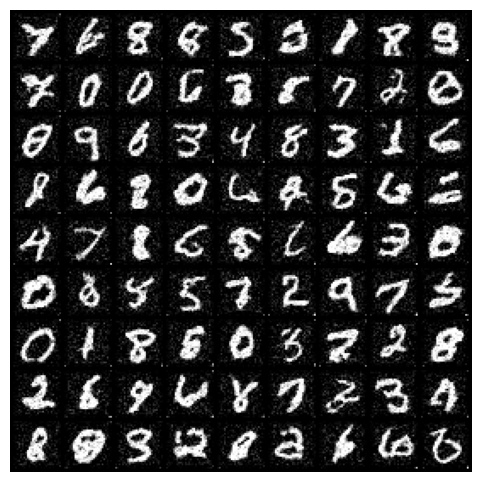}
    \end{minipage}
    
    \vspace{0.2em} 
    
    \begin{minipage}{0.15\textwidth}
        \centering
        \includegraphics[width=\textwidth]{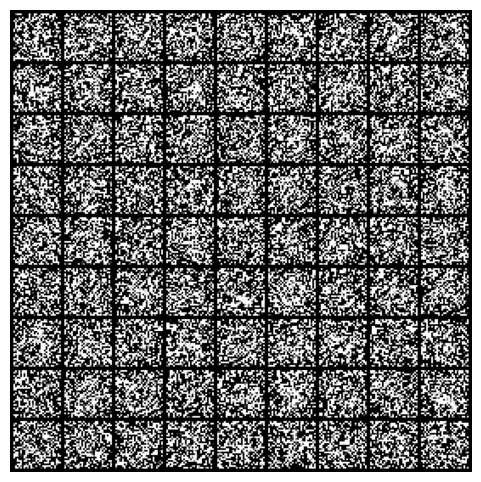}
    \end{minipage}
    \hspace{9em} 
    \begin{minipage}{0.15\textwidth}
        \centering
        \includegraphics[width=\textwidth]{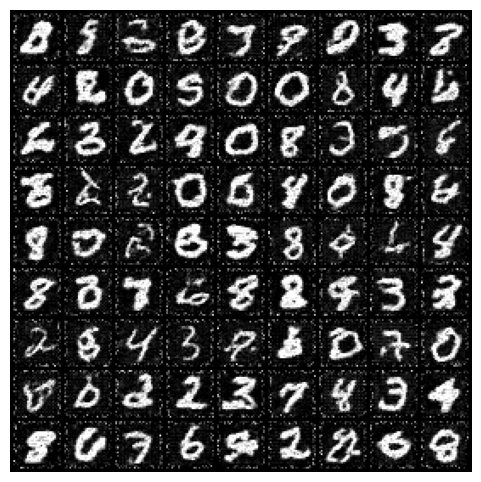}
    \end{minipage}

    
    
    \caption{
    Image generation using the original denoising score matching (left column) versus the regularized version (right column) for different time steps, 
$T=500,T=50$, and $T=20$ (from top to bottom). The middle column displays $81 $ original samples from the \mnist~dataset for comparison with images of dimensions  $\Dim=28\times28\times 1=784$. 
}
    \label{fig:MNIST}
\end{figure*}

We now compare original score matching  and the regularized version on \mnist~dataset~\cite{LeCun1998} including $n=50\,000$ training samples. 
We are interested to time steps $T\in \{500,50,20\}$ for sampling and we consider regularization with $\tuning=0.0005$ for $T=500$ and $\tuning=0.003$ for $T\in \{50,20\}$.
In fact, we set the tuning parameter as a decreasing function of $T$, let say $\tuning=f[T]\in O(c'/T)$ for a real constant $c'\in (0,\infty)$.
Note that two models are already trained over the same amount of data and identical settings employing different objectives. 
For sampling (starting from pure noise), then we try different values of time steps $T\in \{500,50,20\}$. 
That means, for small values of $T$, we just need to pick up a \textbf{larger step size} as we always start from pure noise (see Algorithm~\ref{alg:samp}).
Figure~\ref{fig:MNIST} shows the results. While the original score matching fails to generate reasonable samples for small values of $T$, our proposed score function performs successfully even for $T=20$. 
Let note that generating $64$ samples with $T=500$ steps takes about $11$ seconds, while using $T=50$ steps can reduce runtime to under $1$ second, a tenfold speedup.
In all our simulations, we use the same network structure, optimization method, and sampling approach with identical settings for both approaches (see Appendix~\ref{sec:netstr} for detailed settings). The only difference lies in the objective functions: one is regularized, while the other is not. While it could be argued that alternative network structures or sampling processes might enhance the quality of the generated images for original score matching, our focus remains on the core idea of regularization fixing all other factors and structures.  
We defer the enhanced versions of our simulations aimed at achieving higher-quality images to future works.

\subsection{FashionMNIST}\label{sec:fmnist}
We follow almost all the settings as in Section~\ref{sec:mnist} with $\tuning=0.0001$ for $T=500$ and $\tuning=0.002$ for $T\in \{70,50\}$ for \fmnist~dataset~\cite{Xiao2017} including $n=50\,000$ training samples.  
Results are deferred to Figure~\ref{fig:FMNIST} in the Appendix.
Following the generated images obtained using both approaches, and ensuring that all factors except the objective functions remain identical, we observe that the original score-matching approach produces samples that appear oversmoothed and exhibit imbalanced distributions (see the first image of the left panel of Figure~\ref{fig:FMNIST}).  
In contrast, our regularized approach with a considerably small tuning parameter, generates images that resemble the true data more closely and exhibit a more balanced distribution (see the first image of the right panel of Figure~\ref{fig:FMNIST}). 
For instance, the percentages of generated images for Sandals, Trousers, Dresses, Ankle Boots, and Bags are approximately $(0.0, 0.7, 2.0, 3.0, 4.0)$ using the original score matching, compared to $(8.0, 6.0, 8.0, 10.0, 10.0)$ with the regularized version, highlighting the clear imbalance in distribution for the original score matching.

%% file: Contents/conclusion.tex
Our mathematical proofs (Section~\ref{Sec:main}) and empirical illustrations (Section~\ref{sec:sim}) demonstrate that regularization can reduce the computational complexity of diffusion models considerably. 
Broadly speaking,
regularization replaces the dependence on the input dimension by a dependence on a much smaller intrinsic dimension.
But our findings might just be the beginning: we believe that types of regularization beyond the sparsity-inducing $\ell_1$-regularization applied here, such as total variation, could lead to further improvements.
Finally, exploring sparsity in more structured domains—such as wavelet or Fourier bases, where many signals are naturally sparse—offers a promising and interpretable direction for future work.


%% file: Contents/Further-Exp.tex
We provide additional simulation supporting our theories on further image dataset  Butterflies in Section~\ref{sec:BF} and CIFAR10 in Section~\ref{sec:CIFAR}. 
We then provide a detailed discussion over Assumption~\ref{Assum:approxs} in Section~\ref{sec:assumdisc} including sparsity within scores.
We   introduce our training and sampling approach in Section~\ref{sec:TrainingAlg} and provide details about network architecture and training settings in Section~\ref{sec:netstr}. 
We also conducted additional simulations to compare the perfomance of our method vs using other
sparsity-inducing regularizers in Section~\ref{sec:grouplasso}.
We finally study runtimes in Section~\ref{sec:runtime}. 


\begin{figure*}[htbp]
    \centering
    \begin{minipage}{0.28\textwidth}
        \centering
        \includegraphics[width=\textwidth]{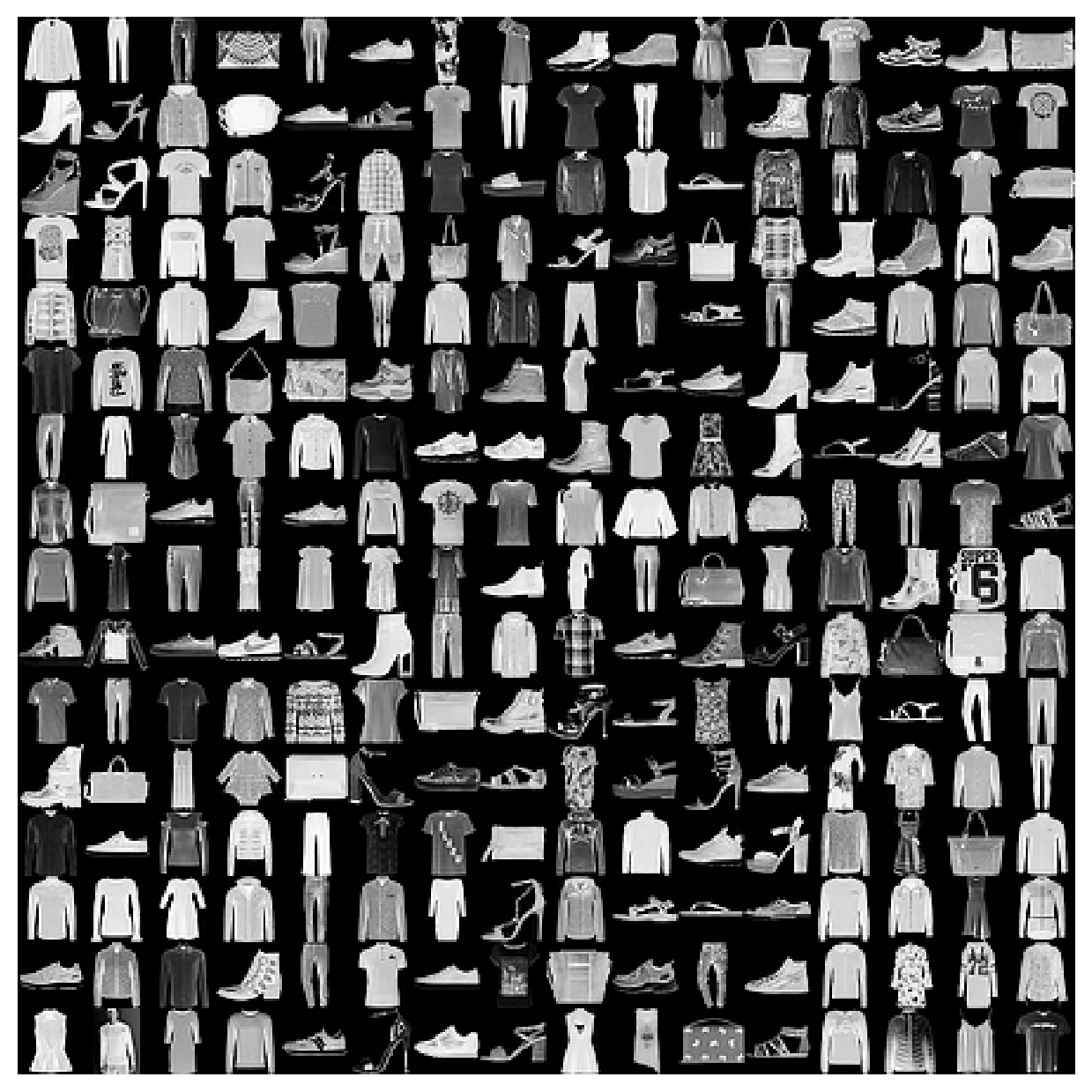}
    \end{minipage}
    
    \vspace{0.1em} 
    \begin{minipage}{0.28\textwidth}
        \centering
        \includegraphics[width=\textwidth]{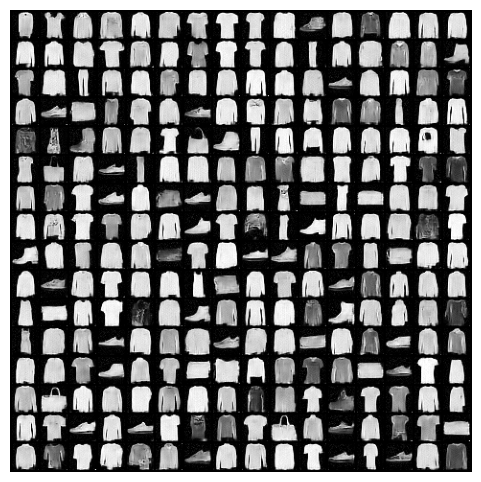}
    \end{minipage}
    \hspace{13em} 
    \begin{minipage}{0.28\textwidth}
        \centering
        \includegraphics[width=\textwidth]{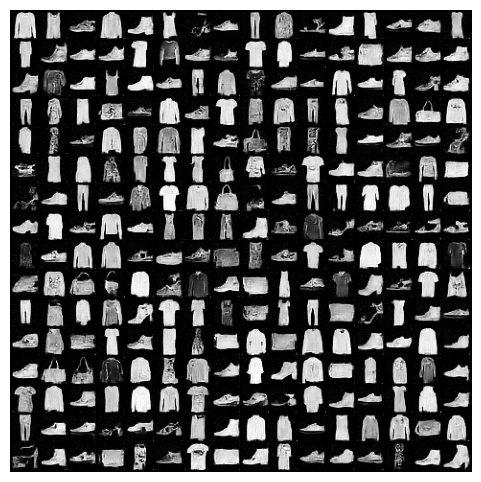}
    \end{minipage}

    \vspace{0.1em} 
    
    \begin{minipage}{0.28\textwidth}
        \centering
        \includegraphics[width=\textwidth]{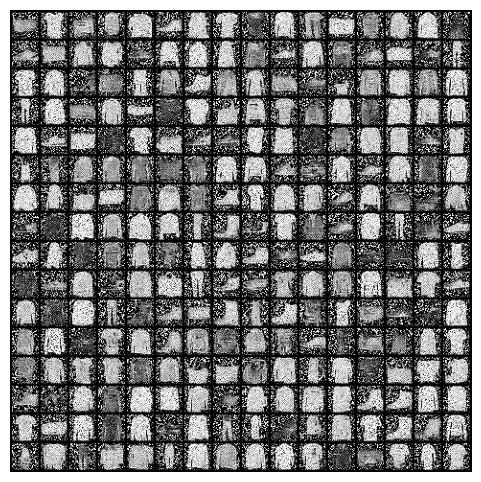}
    \end{minipage}
    \hspace{13em} 
     \begin{minipage}{0.28\textwidth}
        \centering
        \includegraphics[width=\textwidth]{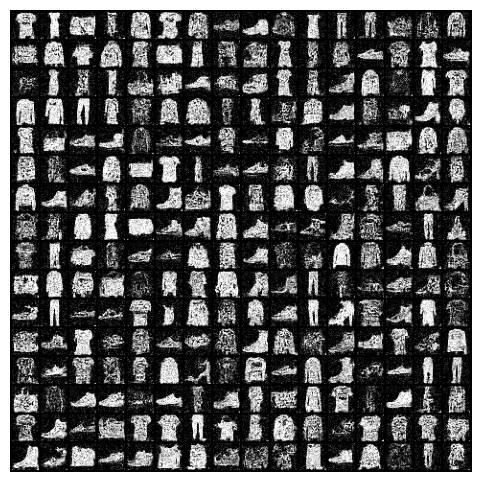}
    \end{minipage}
    
\vspace{0.1em}

    \begin{minipage}{0.28\textwidth}
        \centering
        \includegraphics[width=\textwidth]{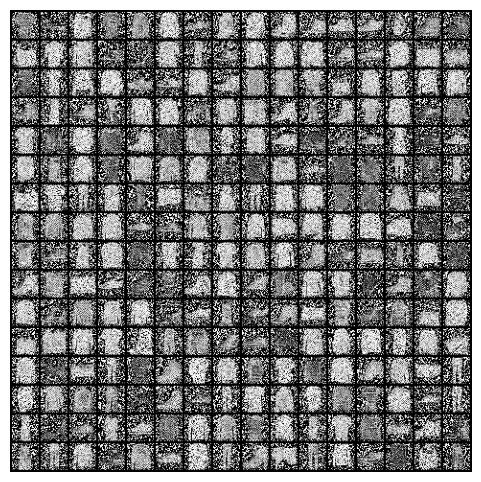}
    \end{minipage}
     \hspace{13em} 
    \begin{minipage}{0.28\textwidth}
        \centering
        \includegraphics[width=\textwidth]{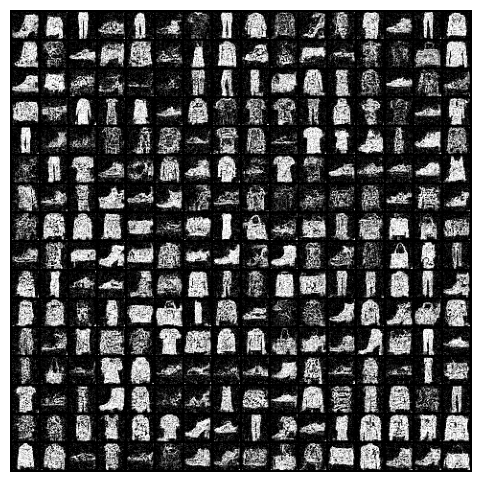}
    \end{minipage}
    
    \caption{Image generation using the original denoising score matching (left column) versus the regularized version (right column) for different time steps,  $T=500, T=70$, and $T=50$ (from top to bottom).  
    The middle column displays $256$ original samples from the
\fmnist~dataset for comparison with images of  dimensions $\Dim=28\times28\times 1=784$.
 Our regularized version generates high-quality images for  
$T=500$ (comparable to the original denoising score matching) and still produces good images even for 
samll $T$, while the original denoising score matching totally fails.
Another notable observation is that our regularization results in more balanced image generation, as evident when comparing our method to the original denoising score matching at $T=500$, where the latter produces overly smooth images.
}\label{fig:FMNIST}
\end{figure*}

\subsection{Butterflies}\label{sec:BF}
\begin{figure*}[htbp]
    \centering
    \begin{minipage}{0.29\textwidth}
        \centering
        \includegraphics[width=\textwidth]{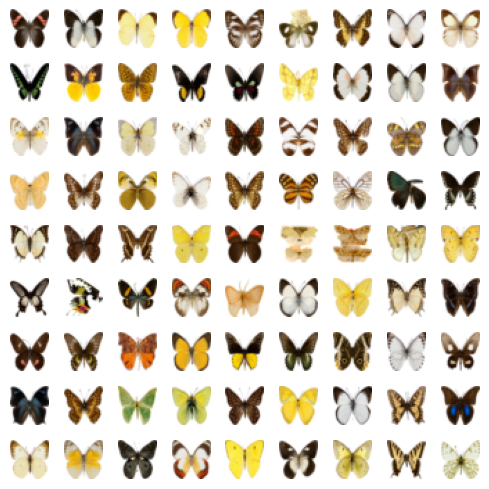}
    \end{minipage}
    
    \vspace{0.1em} 
    \begin{minipage}{0.29\textwidth}
        \centering
        \includegraphics[width=\textwidth]{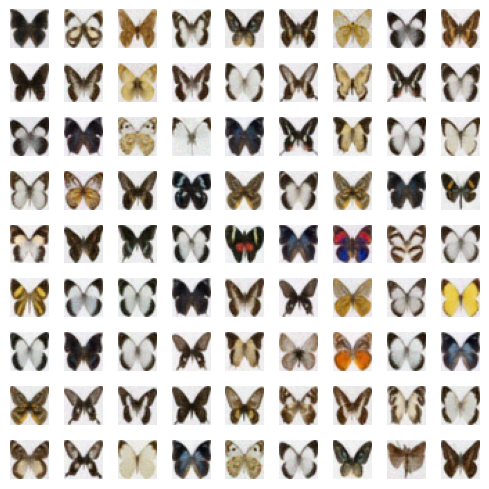}
    \end{minipage}
    \hspace{13em}
     \begin{minipage}{0.29\textwidth}
        \centering      \includegraphics[width=\textwidth]{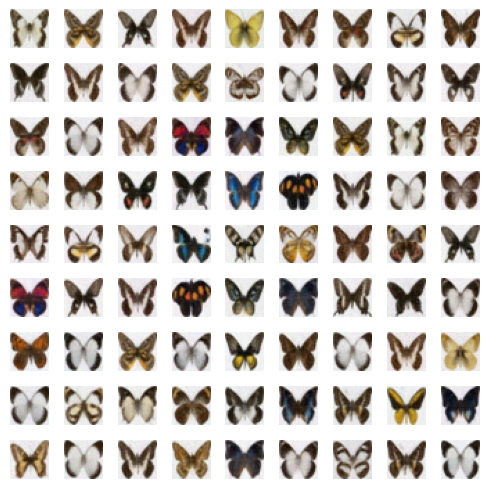}
    \end{minipage}
    
    \vspace{0.1em}
    
    \begin{minipage}{0.29\textwidth}
        \centering
        \includegraphics[width=\textwidth]{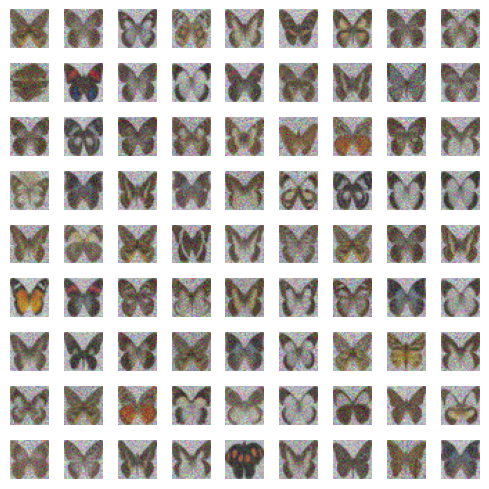}
    \end{minipage}
    \hspace{13em}
    \begin{minipage}{0.29\textwidth}
        \centering
        \includegraphics[width=\textwidth]{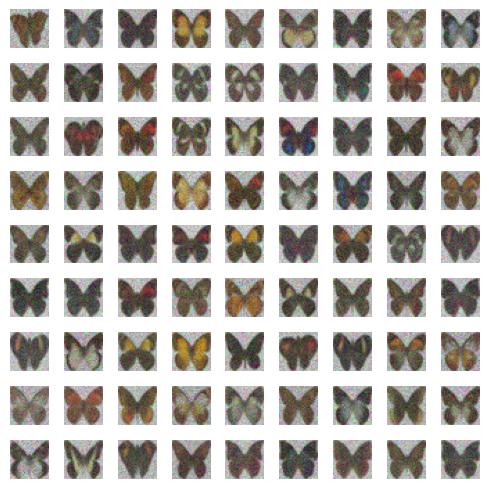}
    \end{minipage}

     \vspace{0.1em} 
    
    \begin{minipage}{0.29\textwidth}
        \centering
        \includegraphics[width=\textwidth]{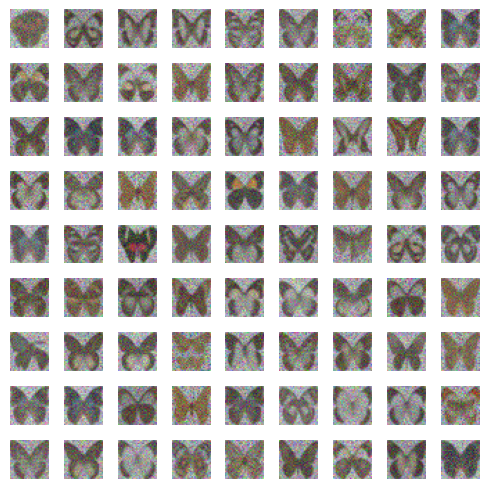}
    \end{minipage}  
    \hspace{13em} 
    \begin{minipage}{0.29\textwidth}
        \centering
        \includegraphics[width=\textwidth]{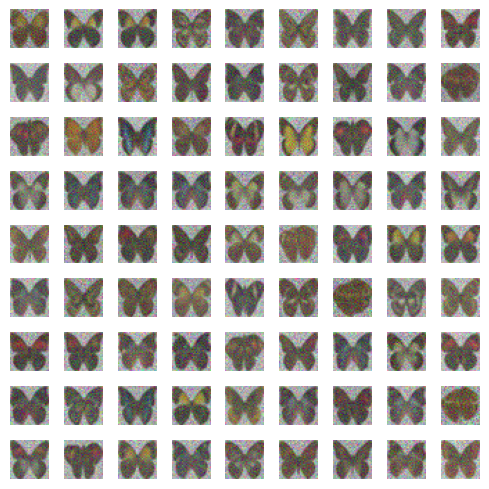}
    \end{minipage}
    
    \caption{Image generation using the original denoising score matching (left column) versus the regularized version (right column) for different time steps,  $T=1000, T=200$, and $T=150$ (from top to bottom).  
    The middle column displays $81$ original samples from the
\ButF~dataset for comparison.
The dataset consists of images with dimensions $\Dim=28\times28\times 3=2352$. As shown in the images, our regularized version  generates high-quality images for  
$T=1000$ (comparable to the original denoising score matching) and still perform better than original denoising score matching for  
$T=200$ and $T=150$.}\label{fig:BF}
\end{figure*}

We also compare original diffusion and  regularized analog on \ButF~dataset (smithsonian-butterflies) including $n=10\,000$ training samples.
We consider regularization $\tuning=0.0001$ for $T=1000$ and $\tuning=0.0005$ for $T\in \{200,150\}$.
Results are provided in Figure~\ref{fig:BF}. 
Again, our results show that our approach perform better than the original score matching  for small values of $T$. 

\subsection{CIFAR10}\label{sec:CIFAR}
\begin{figure*}[htbp]
    \centering
    \begin{minipage}{0.25\textwidth}
        \centering
        \includegraphics[width=\textwidth]{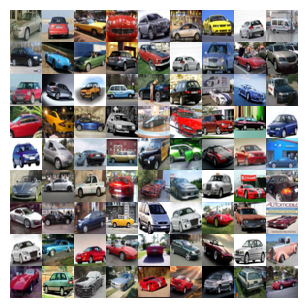}
    \end{minipage}
    
    \vspace{0.1em} 
    \begin{minipage}{0.25\textwidth}
        \centering
        \includegraphics[width=\textwidth]{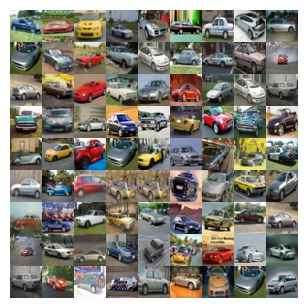}
    \end{minipage}
    \hspace{9em}
     \begin{minipage}{0.25\textwidth}
        \centering
        \includegraphics[width=\textwidth]{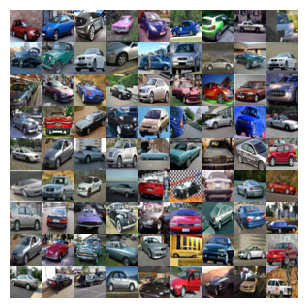}
    \end{minipage}

      \vspace{0.1em}

    
    
    \begin{minipage}{0.25\textwidth}
        \centering
        \includegraphics[width=\textwidth]{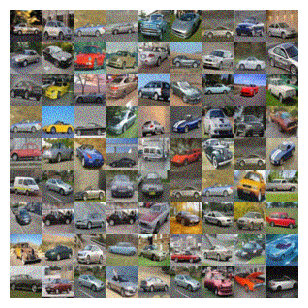}
    \end{minipage}
    \hspace{9em}
    \begin{minipage}{0.25\textwidth}
        \centering
        \includegraphics[width=\textwidth]{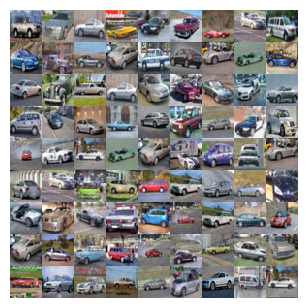}
    \end{minipage}

    \caption{Image generation using the original denoising score matching (left column) versus the regularized version (right column) for different time steps,  $T=1000$ and $T=200$ (from top to bottom).  
    The middle column displays $81$ original samples from the
\cifar~dataset for comparison.
The dataset consists of images with dimensions $\Dim=32\times32\times 3=3072$. As shown in the images, our regularized version with $\tuning=0.0001$ generates high-quality images for  
$T=1000$ and still performs better than original score matching for 
$T=200$.}\label{fig:CIFAR-L}
\end{figure*}

We  compare original diffusion and  regularized analog on \cifar~dataset with $n=50\,000$ training samples.
We employed a large model with a base channel size of $128$. For clarity of comparison, we  report a collection of generated car images in Figure~\ref{fig:CIFAR-L}. Moreover, we computed the \emph{Fréchet Inception Distance} ($\operatorname{FID}$) scores for both models. The original score-matching setup achieves an $\operatorname{FID}$ of $25$, consistent with the results reported in~\cite{song2019generative}. While  we were able to further improve the $\operatorname{FID}$ to $23$ employing our regularized objective with $\tuning=1e-4$.
For $T=500$, our regularized approach ends with $\operatorname{FID}=25$, while original score matching drops to $38$.
For small time budget, $T=200$, the original score matching drops in FID to $160$, while our approach ends with $\operatorname{FID}=49$ for $\tuning=5e-4$.

\subsection{Discussion over Assumption~\ref{Assum:approxs}}  \label{sec:assumdisc} We consider a worst-case scenario for our estimator and examine the validity of Assumption~\ref{Assum:approxs}. 
According to the statement of Theorem~\ref{the:maindS}, let set the tuning parameter as $\tuning=\ror =\max (1/T,\sqrt{\log (np)/ n})$, where we omit constants for simplicity here. Then, the objective function seeks a pair $(\hat{\kappa}, \hat{\Theta})$ such that \begin{align*} (\hat{\kappa}, \hat{\Theta}) \in \arg\min_{\kappa, \Theta \in \paramspaceone} \frac{1}{T} \sum_{t=1}^{T} \left( \frac{1}{n} \sum_{i=1}^{n} \left| \kappa s_{\Theta}(x_t^i, t) - \nabla \log q_t(x_t^i) \right|^2 \right) + \ror \kappa^2. \end{align*}

Suppose $(\kappa_o, \Theta_o)$ is the estimator from the original score matching, i.e., it perfectly fits the true score function. 
Then, \begin{align*} \frac{1}{T} \sum_{t=1}^{T} \left( \frac{1}{n} \sum_{i=1}^{n} \left| \kappa_o s_{\Theta_o}(x_t^i, t) - \nabla \log q_t(x_t^i) \right|^2 \right) + \ror\kappa_o^2 \approx 0.0 + \ror \kappa_o^2. \end{align*}

While our estimator may still yield a smaller objective value compared to the original score matching implying 
\begin{align*} \inf_{\kappa, \Theta \in \paramspaceone} \frac{1}{T} \sum_{t=1}^{T} \left( \frac{1}{n} \sum_{i=1}^{n} \left| \kappa s_{\theta}(x_t^i, t) - \nabla \log q_t(x_t^i) \right|^2 \right) + \ror \kappa^2 \le \ror\kappa_o^2\,. \end{align*} 

Therefore, for  our estimator $(\DnSMR,\scaleM)$ we have 
\begin{align*} \frac{1}{T} \sum_{t=1}^{T} \left( \frac{1}{n} \sum_{i=1}^{n} \left| \scaleM s_{\DnSMR}(x_t^i, t) - \nabla \log q_t(x_t^i) \right|^2 \right) + \ror\scaleM^2 \le \ror \kappa_o^2 
\end{align*} that implies $\scaleM < \kappa_o$.
Note that the scale term reflects the sparsity level since it determines the radius of the optimal unit ball. Despite this reduced scale, the estimator still achieves small score estimation error: \begin{align*} \frac{1}{T} \sum_{t=1}^{T} \left( \frac{1}{n} \sum_{i=1}^{n} \left| \scaleM s_{\DnSMR}(x_t^i, t) - \nabla \log q_t(x_t^i) \right|^2 \right)< \ror \kappa_o^2~\le~\max\Bigl(\frac{1}{T},\frac{\sqrt{\log (np)}}{\sqrt{n}}\Bigr)\kappa_o^2\,.
\end{align*}

These last two points show that our regularized estimator either performs comparably to the original score matching estimator (if no better $(\scaleM, \DnSMR)$ exists), or it finds a better pair with smaller scale and only a small estimation  error for score functions. 
In such cases, we may interpret $\scaleM s_{\DnSMR}(\x_t^i, t)$ as a sparse estimator for $\nabla \log q_t$ in Assumption~\ref{Assum:approxs}, since $\scaleM < \kappa_o$, indicating a smaller radius for the unit ball, which gives a small estimation error for scores. Although Assumption~\ref{Assum:approxs} is stated for the expected error, not the in-sample one, this still gives a meaningful interpretation in practice. We would also like to highlight the expressive power of the network space $\kappa s_{\Theta}(\cdot)$ with $\Theta \in \mathcal{B}_1$, which, following the original work by~\cite{taheri2021}, can approximate the entire network space $s_{\Omega}$ under some assumptions, where $\Omega$ denotes an unconstrained parameter space. This is related to how interpret $\scaleM s_{\DnSMR}(.)$ as $\nabla \log q^s(.)$.
Our discussion above shows that our estimator may behave, in the worst case, similarly to the original score matching. However, it can perform significantly better in the sense that it induces sparsity (by finding a smaller scale $\scaleM \le \kappa_{o}$) while still achieving a  small estimation error for the score of the order of $\max(1/T,1/\sqrt{n})\scaleM^2$ (note that $\scaleM\approx sB$).

Also, as an extreme and simple case of sparse scores one can consider a two-dimensional dataset, where the distribution along the \( x \)-axis is Gaussian, while along the \( y \)-axis it is uniform  and \( x \) and \( y \) are independent. The score function is sparse, given by \( \nabla \log p(x, y) = (\partial_x \log p(x), 0) \), indicating no contribution from the \( y \)-direction.
This toy example can be naturally extended to more complex datasets, such as images, where includes local regularity like certain regions (e.g., background pixels) exhibit near-uniform distributions and thus contribute negligibly to the score. Such structure leads to sparsity in the score function, which is consistent with our assumption, and in fact our empirical observations already well support that.

\paragraph{Assumption~\ref{Assum:approxs} on toy  data} We conducted an experiment to train a diffusion model for generating mixed Gaussian--Uniform data in a $d$-dimensional space. 
We varied the number of Gaussian features $s$, with the remaining $d-s$ dimensions drawn from a uniform distribution independent of the other features.
For each setting, we trained a diffusion model and evaluated the quality of the generated samples by computing the KL divergence between the generated and original distributions.
Results show that our regularized approach, using a fixed tuning parameter $\lambda=0.001$, not only outperforms standard score matching when the sparsity level is high (i.e., many features are uniform), but also consistently dominates standard score matching even when the sparsity level is low. 

\paragraph{Assumption~\ref{Assum:approxs} on MNIST} We conducted an experiment on MNIST to validate the sparsity assumption for a diffusion model trained with standard score matching. 
We measured score approximation error across all time steps for sparsity levels $\{600, 400, 200, 100, 20\}$, obtaining overall errors of approximately $\{0.05, 0.09, 0.18, 0.30, 1.0\}$. 
Errors were also computed separately for early and late time steps: early steps gave $\{0.01, 0.05, 0.10, 0.17, 0.61\}$, while late steps gave $\{0.32, 0.46, 0.89, 2.9, 5.1\}$. 
Since early steps dominate in diffusion models, these results indicate that the sparsity assumption holds well in practice, with negligible bias up to moderate sparsity levels. 
Furthermore, MNIST image generation with small $T$ shows even smaller bias than suggested by the above errors, likely due to score estimation and optimization effects, and our method performs well even for high sparsity.

We would note  that Assumption~\ref{Assum:approxs} can also  be relaxed as: 
\begin{assumption}[Relaxed-sparsity]\label{ass:Rsparsity}
There is a sparsity level $s\in\{1,2,\dots\}$ and an accuracy $\epsilon\in(0,\infty)$, 
$\epsilon\leq1/s^{a}T^{a'}$ ($a',a'\in (0,\infty)$),
such that for all $t \in \{1,\dots,T\}$, there is an analytic auxiliary function $\Fdds_t(\x_t)$ and the corresponding score 
$\nabla_{\x_t}\log\Fdd_t^s(\x_t)$ that is $s$-sparse and $\epsilon$-accurate:
\begin{multline*}
\E_{X_t \sim \Fd_t}\norm{\nabla_{X_t}\log\Fdd_t^s(X_t)}_0~\leq~s~~\text{and} \\
\frac{1}{T} \sum_{t=1}^T
   \sqrt{\E_{X_t \sim \Fd_t} \norm{\nabla_{\x_t}\log\Fdd_t(X_t) - \nabla_{\x_t}\!\log\Fdds_t(X_t)}^2}~\leq~\Esparsity\,.
\end{multline*}
\end{assumption}
Collecting the arguments above, we conclude that by assuming sparsity in the score functions, we incur a bias term of order \( 1/s^{a} T^{a'} \); however, this trade-off allows us to improve the overall convergence rates to \( O(s/\sqrt{n}+1/s^a T^{a'}) \), in contrast to the \( O(d/\sqrt{n}) \) rates established in~\cite{zhu2023sample}.
By selecting the tuning parameter appropriately, we ensure that the trade-off between bias and variance is well balanced.


Recently, regularization has been employed to prevent memorization~\cite{gabriel2025kernel,baptista2025memorization}. 
The analysis in~\cite{baptista2025memorization} highlights the necessity of regularization to avoid reproducing the analytically tractable minimizer, and in doing so, lays the groundwork for a principled understanding of how to regularize effectively. 
Their numerical experiments investigate the properties of: (i) Tikhonov regularization; (ii) regularization schemes designed to promote asymptotic consistency; and (iii) implicit regularizations induced either by under-parameterization of a neural network or by early stopping during training.

\begin{figure}
    \centering
    \includegraphics[width=0.5\linewidth]{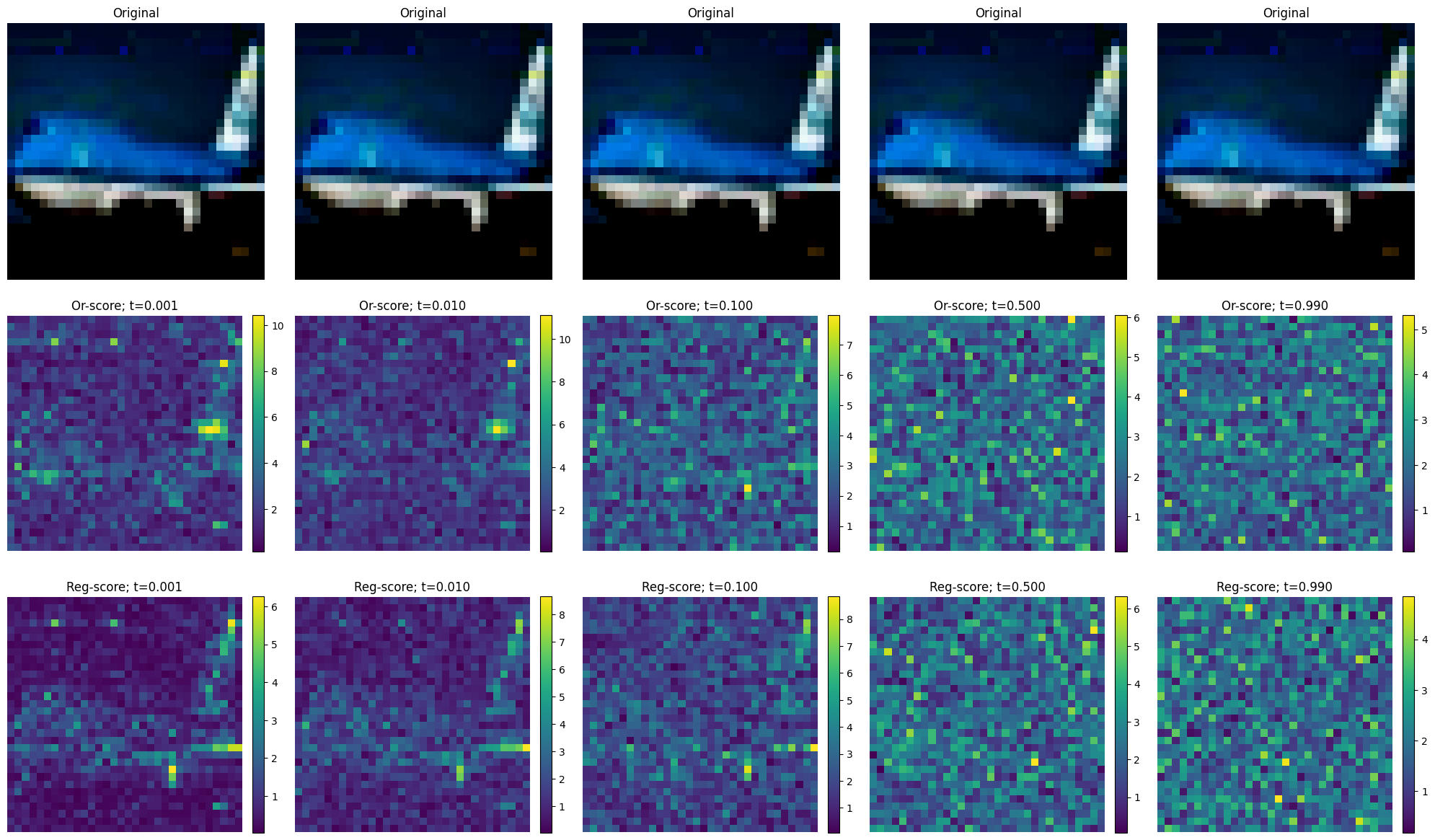}
    \caption{Heat map of pixel-wise score magnitudes, highlighting informative regions where the model assigns high sensitivity.}
    \label{fig:score-heat}
\end{figure}

\paragraph{Sparsity within scores}
We obtain score heat maps by computing the pointwise norm of the estimated score field 
 for each pixel for CIFAR10 in Figure~\ref{fig:score-heat}. This highlights regions where the model assigns high sensitivity, revealing structural or edge-related features. Our simulations reveal two key observations: (1) results exhibit a fraction of near-zero score coefficients, especially at small time steps where the perturbed image retains meaningful structure; 
 (2) this behavior is fully consistent with our earlier discussion of local regularity—such as background regions and near-uniform pixel areas—which naturally contribute very little to the score field; see Figure~\ref{fig:score-heat}, where background pixels exhibit highly sparse scores while structurally important edge pixels remain distinctly non-zero in early time steps.


\subsection{Training and sampling algorithms}\label{sec:TrainingAlg}
Here we provide details about how we solve the objective function~\eqref{dscorematchR} in practice, that is, how we deal with the expected values and score functions. 

Let first define the objective function over a batch  of training examples $\x_{\batchS}$ (a batch of size $\batchS\in \{1,2,\dots\}$) and for a batch of random time steps $\boldsymbol{t}_{\batchS}\in (0,1]^{\batchS}$: 
\begin{equation}\label{eq:simobj}
f(\scale,\Theta,\x_{\batchS},\boldsymbol{t}_{\batchS})\deq \frac{1}{\batchS}\sum_{i=1}^{\batchS}\norm{\scale\scoref(\x^i_{t_i},t_i)-\nabla_{\x_{t}}\log \Fdd_t(\x^i_{t_i}|\x^i)}^2\bigr]
 +\tuning \scale^2 
\end{equation}
with 
\begin{equation}\label{eq:pertx}   \Fd_t(\x^i_{t_i}|\x^i)=\mathcal{N}\bigl(\x^i,\sigma_{t_i}\Identity\bigr) 
\end{equation}
with $\sigma_t\deq (\sigma^{2t}-1)/(2\log \sigma)$ for $t\in (0,1]$ and a large enough $\sigma\in (0,\infty)$ (we set $\sigma=5$ for \ButF~and \cifar~and $\sigma=25$ for other datasets). 
Here $\x^i_t$ corresponds to a perturbed version of the training sample $\x^i$ ($i$th sample of the batch) in  time step $t$. As stated in~\eqref{eq:pertx}, once $\sigma$ is large, $\x_{1}$ ($t=1$) goes to a mean-zero Gaussian. 
And as shown in~\cite{vincent2011connection}, the optimization objective  $\E_{\Fdd_t(\x_t|\x)\Fdd_{0}(\x)}[\norm{\scale\scoref(\x_{t},t)-\nabla_{\x_{t}}\log \Fdd_t(\x_{t}|\x)}^2]$ for a fixed variance~$\varnoise_t$ is equivalent to the optimization  objective $\E_{\Fdd_t(\x_t)}[\norm{\scale\scoref(\x_{t},t)-\nabla_{\x_{t}}\log \Fdd_t(\x_{t})}^2]$ and, therefore, satisfies $\scale^*\scorefS(\x_{t},t)=\nabla_{\x_t}\log \Fdd_t(\x_t)$.
We then provide Algorithm~\ref{alg:training} for solving the objective function in~\eqref{dscorematchR}. 
Note that we can easily compute the score functions in~\eqref{eq:simobj} since there is a  closed-form solution for them as densities are just Gaussian conditional on $\x^i$. 
\begin{algorithm}[H]
\caption{Training algorithm}\label{alg:training}
\begin{algorithmic}[1] 
\STATE \textbf{Inputs:} $\sigma$, $n_{\operatorname{epochs}}$ (number of epochs), $\batchS(\operatorname{batch-size})$, $\operatorname{eps}=0.00001$
\STATE \textbf{Outputs:} $(\DnSMR,\scaleM)$
\STATE Initialize parameters $(\DnSMR,\scaleM)$ 
\FOR{$i = 1$ to $n_{\operatorname{epochs}}$}
\FOR{$\x_{\batchS}  $ in data-loader}
\STATE $\boldsymbol{t}_{\batchS}=\{\mathcal{U}_{[0,1]}\}^{\batchS}  (1 - \operatorname{eps}) + \operatorname{eps}$ 
    \STATE One step optimization minimizing   $f(\scale,\Theta,\x_{\batchS},\boldsymbol{t}_{\batchS})$ in~\eqref{eq:simobj} employing a random batch of time steps $\boldsymbol{t}_{bs}\in (0,1]^{bs}$  and updating $(\DnSMR,\scaleM)$
\ENDFOR
\ENDFOR

\end{algorithmic}
\end{algorithm}
Parameter $\operatorname{eps}$ in Algorithm~\ref{alg:training}  is introduced for numerical stability and to refuse $t=0$. 
For a sufficiently large number of epochs, we expect to learn the scores accurately for different time steps. 
For sampling process, we employ a naive sampler as proposed in Algorithm~\ref{alg:samp}
employing Langevin dynamics~\cite[Section~2.2]{song2019generative} to align with our theory.  
\begin{algorithm}[H]
\caption{Sampling algorithm}\label{alg:samp}
\begin{algorithmic}[1] 

\STATE \textbf{Inputs:} $\sigma$, $\operatorname{eps}=0.00001$, T (Time steps)
\STATE \textbf{Output:} \x
\STATE $\x=\x_{\operatorname{init}}=\StandNormal\sigma_1$ 
\STATE $\boldsymbol{t} = linspace(1., eps, T)$ (make a grid of time steps)
\STATE $\eta = \boldsymbol{t}[0] - \boldsymbol{t}[1]$ (set step size)
\FOR{t in $\boldsymbol{t}$}
\STATE $\x=\x+\eta \scaleM\score_{\DnSMR}(\x,t)+\sqrt{2\eta}\StandNormal$ (update $\x$)
\ENDFOR
\end{algorithmic}
\end{algorithm}
\subsection{Network architecture and training settings}\label{sec:netstr}
Our model is a U-Net architecture with 4 downsampling and 4 upsampling layers, each comprising residual blocks. The network starts with a base width of 32 channels, doubling at each downsampling step to a maximum of 256 channels, and mirrors this in the decoder. A bottleneck layer with 256 channels connects the encoder and decoder. Time information is encoded using Gaussian Fourier projections and injected into each residual block via dense layers. Group normalization is applied within the residual blocks, and channel attention mechanisms are included selectively to enhance feature representations.
For training, we used the Adam optimizer with a learning rate of 0.001, and for sampling, we employed a signal-to-noise ratio of 0.1.
We used a batch size of $128$ and trained for $2000$ epochs on the \ButF~dataset and less than $1000$ epochs on the other datasets.

\subsection{Comparison with other sparsity-inducing regularizers}\label{sec:grouplasso}

We conducted additional  simulations to compare the perfomance of our method vs using other sparsity-inducing regularizers.

\begin{enumerate}
    \item \textbf{$\ell_2$-regularization:} Although the $\ell_2$-norm helps control the growth of magnitudes, it does not promote sparsity by design. Our complementary simulations show that using $\ell_2$-regularization does not significantly speed up inference time in our experiments.
    
    \item \textbf{Group lasso:} Our simulations show that applying group lasso on images—considering groups as $p \times p$ pixel blocks ($4\times 4$)—can also improve inference time and, for some datasets like MNIST, even outperform $\ell_1$-regularization.
    
    \item \textbf{Combination of norms:} Employing a combination of norms, namely group lasso plus $\ell_1$-norm, performs comparably to individual regularizations for the MNIST family, but we do not observe any considerable improvement over using either group lasso or $\ell_1$-norm alone.
\end{enumerate}

\subsection{Runtimes}\label{sec:runtime}
We measured the optimization time for training the regularized and the original diffusion models on MNIST over $50$ epochs using Adam on the same device. The training time for the original DFM is approximately $20$ minutes in total, while the regularized version takes about $21$ minutes.
Both objectives converge after about $20$ epochs. This shows that adding regularization has only a minor effect on the training time. Similar behavior also observed for other datasets.

%% file: Contents/Proofs.tex
\section{Auxiliary results and proofs}
In this section,  we first provide  two auxiliary results and then, we present detailed proofs of our main results.
Note that throughout our proofs, we will omit the index  $\x_t$ from $\nabla_{\x_t}\log \Fdd_t(\cdot)$ for simplicity in notations and simply write $\nabla\log \Fdd_t(\cdot)$.  

\subsection{Auxiliary results}\label{sec:Aux}
We denote the distribution of the latent steps in the reverse process, utilizing the sparse gradient vector, as
 $\Bds_{t-1|t}\deq\mathcal{N}(\x_{t-1};\us_t(\x_t),\sigma_t^2\Identity)$ with 

\begin{equation}\label{eq:mus}
    \us_t(\x_t)\deq\frac{1}{\sqrt{\alpha_t}}\bigl(\x_t+(1-\alpha_t)\nabla_{\x_t}\log \Fdds_t(\x_t)\bigr)\,.
\end{equation}
We then connect the conditional distributions $\Fdd_{t-1|t}(\x_{t-1}|\x_t) $ and $\Bdds_{t-1|t}(\x_{t-1}|\x_t)$ using the following lemma inspired by proof approach of~\cite{liang2024nonN}: 
\begin{lemma}[Tilting factor]\label{lem:tilfactor}
 For a fixed $\x_t$ we have 
 \begin{equation*}
  \Fdd_{t-1|t}(\x_{t-1}|\x_t)  \varpropto \Bdds_{t-1|t}(\x_{t-1}|\x_t)\exp\bigl( \zeta_{t,t-1}(\x_t,\x_{t-1})\bigr)
\end{equation*}
with $\zeta_{t,t-1}(\x_t,\x_{t-1})\deq \log \Fdd_{t-1}(\x_{t-1})-\sqrt{\alpha_t}\x_{t-1}^T\nabla \log \Fdds_t(\x_t)+f(\x_t)$, where $f(\x_t)$ is an arbitrary function of $\x_t$.
\end{lemma}
The notation $\propto$ in the Lemma~\ref{lem:tilfactor} means proportional. 
Lemma~\ref{lem:tilfactor} is employed in the proof of our Lemma~\ref{lem:RevError} and its 
detailed proof is presented in Section~\ref{proof:tilf}.
\begin{lemma}[Log-density of  the backward process]\label{lem:logdeng}
 We have 
 \begin{align*}
     &\sum_{t=1}^T\E_{X_t,X_{t-1}\sim \Fd_{t,t-1}} \biggl[\log\frac{\Bdd^s_{t-1|t}(X_{t-1}|X_t)}{\BddA_{t-1|t}(X_{t-1}|X_t)}\biggr] \\
     &\le  \sum_{t=1}^T (1-\alpha_t)\biggl(\sqrt{\E_{X_t}\norm{\nabla\log \Fdd_t(X_t)-\nabla\log \Fdds_t(X_t)}^2\E_{X_t}\norm{\nabla\log \Fdds_t(X_t)-\scaleh\scorefh(X_t,t)}^2}\\
    &~~~+ \frac{1}{2}\E_{X_t\sim Q_t}\norm{\scaleh\scorefh(X_t,t)-\nabla\log q^s_t(X_t)}^2\biggr)\,.
 \end{align*}   
\end{lemma}
Lemma~\ref{lem:logdeng} is employed in the proof our main Theorem~\ref{the:maindS} and its detailed proof is presented in Section~\ref{proof:loggaussian}.
\subsection{Proof of Theorem~\ref{the:maindS}}\label{proof:mainR}
\begin{proof}

The poof approach is based on decomposing the total error using the Markov property of the forward and the backward process. 
We then, employ our Lemma~\ref{lem:RevError} and Lemma~\ref{lem:emp} to handle individual terms.

Following the proof approach proposed by~\cite[Equation~13]{liang2024nonN} (based on Markov property of the forward and the backward process), we can decompose the total  error as  
\begin{equation*}
    \KL(\Fd_0 ||\BdA_0) \le \KL(\Fd_T||\BdA_T)+\sum_{t=1}^{T} \E_{X_t\sim \Fd_t}\bigl[\KL\bigl(\Fd_{t-1|t}(.|X_t)||\BdA_{t-1|t}(.|X_t)\bigr)\bigr]\,.
\end{equation*}
We then employ the auxiliary function $\Fdds_t(\cdot)$ that  satisfies our Assumption~\ref{Assum:approxs} (it doesn't need to be necessarily unique). 
We also use the notation $\Bdd^s_t(\cdot)$ for the  reverse counterpart  of the auxiliary function $\Fdds_t(\cdot)$ (see~\eqref{eq:mus}).
 We then  rewrite the previous display employing $\Bdd^s_t(\cdot)$ and the definition of $\KL$:
\begin{align*}
    \KL(\Fd_0 ||\BdA_0) &\le \KL(\Fd_T||\BdA_T)+\sum_{t=1}^{T} \E_{X_t\sim \Fd_t}\bigl[\KL\bigl(\Fd_{t-1|t}(.|X_t)||\BdA_{t-1|t}(.|X_t)\bigr)\bigr]\\
&=\underbrace{\KL(\Fd_T||\BdA_T)}_{\text{Term~1: Initialization error}}
+\underbrace{\sum_{t=1}^{T} \E_{X_t,X_{t-1}\sim \Fd_{t,t-1}}\biggl[\log\frac{\Fdd_{t-1|t}(X_{t-1}|X_t)}{\Bdd^s_{t-1|t}(X_{t-1}|X_t)}\biggr]}_{\text{Term~2: Reverse-step error}}\\
&~~~~+\underbrace{\sum_{t=1}^{T} \E_{X_t,X_{t-1}\sim \Fd_{t,t-1}}\biggl[\log\frac{\Bdd^s_{t-1|t}(X_{t-1}|X_t)}{\BddA_{t-1|t}(X_{t-1}|X_t)}\biggr]}_{\text{Term~3: Estimation error}}\,.
\end{align*}
We now need to study each term specified above individually.

\emph{Term~1: Initialization error}

Under the Assumption~\ref{Assum:FSM} and the assumed step size in~\eqref{eq:stepsize}, we have 
with~\cite[Remark~1]{liang2024nonN} 
\begin{align*}
  \KL(\Fd_T||\BdA_T) \le  \frac{\FSM}{T^2}\,,
\end{align*}
As stated by~\cite[Definition~1]{liang2024non}, there exist cases that verify our step-size assumption, which in turn implies that $\Bar{\alpha}_T\le c/T^2$ for a constant $c\in(0,\infty)$. 


\emph{Term~2: Reverse-step error}

Under our Assumption~\ref{Assum:approxs} and Assumption~\ref{assum:ReDe}
and with Lemma~\ref{lem:RevError} we have  
\begin{align*}
    \sum_{t=1}^{T}~&\E_{X_t,X_{t-1}\sim \Fd_{t,t-1}}\biggl[\log\frac{\Fdd_{t-1|t}(X_{t-1}|X_t)}{\Bdd^s_{t-1|t}(X_{t-1}|X_t)}\biggr]\notag\\
    &\le  \frac{1}{T}\bigl(s^2\DerBound^2+  s^2\DerBound^2+ s^2\DerBound^2\epsilon+s\DerBound \epsilon\bigr)+\deltaT\,.
\end{align*}

\emph{Term~3: Estimation error}

We employ 1.~Lemma~\ref{lem:logdeng}, 2.~adding a zero-valued term, 3.~triangle inequality, and 4.~the definition of our estimator and some linear algebra to obtain 
\allowdisplaybreaks
\begin{align*}
    \sum_{t=1}^{T}~&\E_{X_t,X_{t-1}\sim \Fd_{t,t-1}}\biggl[\log\frac{\Bdd^s_{t-1|t}(X_{t-1}|X_t)}{\BddA_{t-1|t}(X_{t-1}|X_t)}\biggr]\\
    &\le  \sum_{t=1}^{T} \frac{1-\alpha_t}{2}  \E_{X_t\sim Q_t}\norm{\scaleh\scorefh(X_t,t)-\nabla\log q^s_t(X_t)}^2\\
    &~~~+\sum_{t=1}^T (1-\alpha_t)\sqrt{\E_{X_t}\norm{\nabla\log \Fdd_t(X_t)-\nabla\log \Fdds_t(X_t)}^2\E_{X_t}\norm{\nabla\log \Fdds_t(X_t)-\scaleh\scorefh(X_t,t)}^2}\\
     &=   \sum_{t=1}^{T} \frac{1-\alpha_t}{2} \biggl(\frac{1}{n} \sum_{i=1}^{n}  \norm{\scaleh\scorefh(\x_t^i,t)-\nabla\log q^s_t(\x_t^i)}^2+\E_{X_t}\norm{\scaleh\scorefh(X_t,t)-\nabla\log q^s_t(X_t)}^2\\
     &~~~-\frac{1}{n} \sum_{i=1}^{n}  \norm{\scaleh\scorefh(\x_t^i,t)-\nabla\log q^s_t(\x_t^i)}^2\biggr)\\
     &~~~+\sum_{t=1}^T (1-\alpha_t)\sqrt{\E_{X_t}\norm{\nabla\log \Fdd_t(X_t)-\nabla\log \Fdds_t(X_t)}^2\E_{X_t}\norm{\nabla\log \Fdds_t(X_t)-\scaleh\scorefh(X_t,t)}^2}\\
     &\le   \sum_{t=1}^{T} \frac{1-\alpha_t}{2} \biggl(\frac{1}{n} \sum_{i=1}^{n}  \norm{\scaleh\scorefh(\x_t^i,t)-\nabla\log q_t(\x_t^i)}^2+\frac{1}{n} \sum_{i=1}^{n}  \norm{\nabla \log q_t(\x_t^i)-\nabla\log q^s_t(\x_t^i)}^2\\
     &~~+\E_{X_t}\norm{\scaleh\scorefh(X_t,t)-\nabla\log q^s_t(X_t)}^2
     -\frac{1}{n} \sum_{i=1}^{n}  \norm{\scaleh\scorefh(\x_t^i,t)-\nabla\log q^s_t(\x_t^i)}^2\biggr)\\
     &~~~+ \sum_{t=1}^T (1-\alpha_t)\sqrt{\E_{X_t}\norm{\nabla\log \Fdd_t(X_t)-\nabla\log \Fdds_t(X_t)}^2\E_{X_t}\norm{\nabla\log \Fdds_t(X_t)-\scaleh\scorefh(X_t,t)}^2}\\
      &\le \sum_{t=1}^{T} \frac{1-\alpha_t}{2}  \biggl(\frac{1}{n} \sum_{i=1}^{n}  \norm{\scale\scoref(\x_t^i,t)-\nabla\log q_t(\x_t^i)}^2+\tuning\scale^2-\tuning\scaleh^2\\
      &~~~+\frac{1}{n} \sum_{i=1}^{n}  \norm{\nabla \log q_t(\x_t^i)-\nabla\log q^s_t(\x_t^i)}^2\\
     &~~~+\Bigl|\E_{X_t}\norm{\scaleh\scorefh(X_t,t)-\nabla\log q^s_t(X_t)}^2
     -\frac{1}{n} \sum_{i=1}^{n}  \norm{\scaleh\scorefh(\x_t^i,t)-\nabla\log q^s_t(\x_t^i)}^2\Bigr|\biggr)\\
     &~~~+ \sum_{t=1}^T (1-\alpha_t) \sqrt{\E_{X_t}\norm{\nabla\log \Fdd_t(X_t)-\nabla\log \Fdds_t(X_t)}^2\E_{X_t}\norm{\nabla\log \Fdds_t(X_t)-\scaleh\scorefh(X_t,t)}^2}
\end{align*}
for arbitrary function $\scoref(\cdot,\cdot)$ with $\Theta\in \paramspaceone$ and $\scale\in (0,\infty)$.

Now, by collecting all the pieces of the proof we obtain 
\begin{align*}
     \KL(\Fd_0 ||\BdA_0) &\le \frac{\FSM}{T^2}+\frac{1}{T}\bigl(s^2\DerBound^2+  s^2\DerBound^2+ s^2\DerBound^2\epsilon+s\DerBound \epsilon\bigr)+\deltaT \notag\\
     &~+ \sum_{t=1}^{T} \frac{1-\alpha_t}{2} \biggl(\frac{1}{n} \sum_{i=1}^{n}  \norm{\scale\scoref(\x_t^i,t)-\nabla\log q_t(\x_t^i)}^2+\tuning\scale^2-\tuning\scaleh^2\\
      &~~+\frac{1}{n} \sum_{i=1}^{n}  \norm{\nabla \log q_t(\x_t^i)-\nabla\log q^s_t(\x_t^i)}^2\\
     &~+\sup_{\NetP\in \paramspaceone}\Bigl|\E_{X_t}\norm{\scaleh\scoref(X_t,t)-\nabla\log q^s_t(X_t)}^2
     -\frac{1}{n} \sum_{i=1}^{n}  \norm{\scaleh\scoref(\x_t^i,t)-\nabla\log q^s_t(\x_t^i)}^2\Bigr|\\
     &~+2\sqrt{\E_{X_t}\norm{\nabla\log \Fdd_t(X_t)-\nabla\log \Fdds_t(X_t)}^2\E_{X_t\sim Q_t}\norm{\nabla\log \Fdds_t(X_t)-\scaleh\scorefh(X_t,t)}^2}\biggr)\,.
\end{align*}
A high level idea now is choosing the tuning parameter  $\tuning$ in such a way that the term $-\tuning\scaleh^2$ can dominate the  terms in the absolute value that are dependent over $\scaleh$. 
The point here is that the terms in the absolute value are growing in the sparse function space.
Employing Lemma~\ref{lem:emp}  we obtain for 
$\tuning \ge  C_{\x} \sqrt{\log (np) /n}$  that 

\begin{align*}
     \KL(\Fd_0 ||\BdA_0) &\le \frac{\FSM}{T^2}+\frac{1}{T}\bigl(s^2\DerBound^2+  s^2\DerBound^2+ s^2\DerBound^2\epsilon+s\DerBound \epsilon\bigr)+\deltaT\notag\\
     &~~~+\inf_{\NetP\in \paramspaceone; \scale\in (0,\infty)}\biggl\{  \sum_{t=1}^{T} \frac{1-\alpha_t}{2}  \frac{1}{n} \sum_{i=1}^{n}  \norm{\scale\scoref(\x_t^i,t)-\nabla\log q_t(\x_t^i)}^2+r\scale^2\biggr\}\\
      &~~~+   \sum_{t=1}^{T} \frac{1-\alpha_t}{2}\biggl(\frac{1}{n} \sum_{i=1}^{n}  \norm{\nabla \log q_t(\x_t^i)-\nabla\log q^s_t(\x_t^i)}^2+ C_x s^2\DerBound^2\frac{\sqrt{\log (np)}}{\sqrt{n}}\\
      &~~~+2\sqrt{\E_{X_t}\norm{\nabla\log \Fdd_t(X_t)-\nabla\log \Fdds_t(X_t)}^2\E_{X_t}\norm{\nabla\log \Fdds_t(X_t)-\scaleh\scorefh(X_t,t)}^2}\biggr)
\end{align*}
with probability at least $1-32/n$.

Using  Assumption~\ref{Assum:approxs},~\eqref{eq:stepsize}, and \eqref{eq:paramspace} we also obtain 
\begin{align*}
     \sum_{t=1}^{T} (1-\alpha_t)&~\sqrt{\E_{X_t\sim Q_t}\norm{\nabla\log \Fdd_t(X_t)-\nabla\log \Fdds_t(X_t)}^2\E_{X_t\sim Q_t}\norm{\nabla\log \Fdds_t(X_t)-\scaleh\scorefh(X_t,t)}^2} \\
   &\le  \Bigl(\max_{t\in \{1,\dots,T\}} \sqrt{\E_{X_t\sim Q_t}\norm{\nabla\log \Fdds_t(X_t)-\scaleh\scorefh(X_t,t)}^2}\Bigr)\\
   &~~~~~\sum_{t=1}^{T} (1-\alpha_t)  \sqrt{\E_{X_t\sim Q_t}\norm{\nabla\log \Fdd_t(X_t)-\nabla\log \Fdds_t(X_t)}^2}\\
  &\le \Esparsity \Bigl(\max_{t\in \{1,\dots,T\}} \sqrt{\E_{X_t\sim Q_t} \bigl(2\norm{\nabla\log \Fdds_t(X_t)}^2+2\norm{\scaleh\scorefh(X_t,t)}^2\bigr)}\Bigr)\\
   &\le \Esparsity\bigl(\sqrt{2(s\DerBound)^2+2\scaleh^2}\bigr) \\
   &\le 5\Esparsity s\DerBound\\
   &\le 5 \frac{s\DerBound}{T}\,,
\end{align*}
where for the fourth inequality  we can follow the same approach as in~\cite[Page~155]{taheri2021} to conclude that for large enough tuning (just double it), $\scaleh\le 3\scaleS$ (where $\scaleS\approx s\DerBound$) that gives us the space to remove $\scaleh$ from our bounds. 
Let's also note that under the Assumption~\ref{Assum:approxs} and \eqref{eq:stepsize}, we can also conclude that 
\begin{align*}
     \sum_{t=1}^{T} (1-\alpha_t) \frac{1}{n} \sum_{i=1}^{n}  \norm{\nabla \log q_t(\x_t^i)-\nabla\log q^s_t(\x_t^i)}^2 \lesssim \Esparsity \le \frac{1}{T}\,.
\end{align*}
Finally, collecting   displays above, some simplifications, and  keeping the dominant factors gives us the desired results
\begin{align*}
     \KL(\Fd_0 ||\BdA_0) &\le \frac{\FSM}{T^2}+\frac{1}{T}\bigl(s^2\DerBound^2+  s^2\DerBound^2+ s^2\DerBound^2\epsilon+s\DerBound \epsilon\bigr)+\deltaT\notag\\
     &~~~+\inf_{\NetP\in \paramspaceone; \scale\in (0,\infty)}\biggl\{ \sum_{t=1}^{T} \frac{1-\alpha_t}{2}   \frac{1}{n} \sum_{i=1}^{n}  \norm{\scale\scoref(\x_t^i,t)-\nabla\log q_t(\x_t^i)}^2+r\scale^2\biggr\}\\
      &~~~+\frac{1}{T}+ s^2\DerBound^2\frac{\sqrt{\log (np)}}{\sqrt{n}}+ 5\frac{s\DerBound}{T}\\
      &\le \frac{\FSM}{T^2}+\frac{1}{T}\max\{1,9(s\DerBound)^2\} + C_{\x}s^2\DerBound^2\frac{\sqrt{\log (nTp)}}{\sqrt{n}}
      +\Delta_{T} (\log \Fdd,\log \Fdds)\\
     &~~~~~~+\inf_{\substack{\NetP\in \paramspaceone\\ \scale\in (0,\infty)}}\biggl\{\frac{\log T}{T}  \sum_{t=1}^{T}\frac{1}{n} \sum_{i=1}^{n}  \norm{\scale\scoref(\x_t^i,t)-\nabla_{\x_t}\!\log\Fdd_t(\x_t^i)}^2
      +\tuning\scale^2\biggr\}\,.
\end{align*}
\end{proof}

\subsection{Proof of Lemma~\ref{lem:RevError}}\label{proof:lemReverror}
\begin{proof}

We start the proof with Lemma~\ref{lem:tilfactor} that relates $\Fdd_{t-1|t}(\x_{t-1}|\x_t)$ and $\Bdd^s_{t-1|t}(\x_{t-1}|\x_t)$  by
\begin{align}
  \Fdd_{t-1|t}(\x_{t-1}|\x_t)  \varpropto \Bdds_{t-1|t}(\x_{t-1}|\x_t)\exp\bigl( \zeta_{t,t-1}(\x_t,\x_{t-1})\bigr)
\end{align}
with $\zeta_{t,t-1}(\x_t,\x_{t-1})= \log \Fdd_{t-1}(\x_{t-1})-\sqrt{\alpha_t}\x_{t-1}^T\nabla \log \Fdds_t(\x_t)+f(\x_t)$, where $f(\x_t)$ is an arbitrary function of $\x_t$.
Now, let's progress with adding a zero-valued term to the $\zeta_{t,t-1}(\x_t,\x_{t-1})$ 

\begin{align*}
\zeta_{t,t-1}(\x_t,\x_{t-1})&=\log \Fdds_{t-1}(\x_{t-1})-\sqrt{\alpha_t}\x_{t-1}^T\nabla \log \Fdds_t(\x_t)\\
&~~~+ \log \Fdd_{t-1}(\x_{t-1})-\log \Fdds_{t-1}(\x_{t-1})+f(\x_t)
\end{align*}
and set $f(\x_t)=-\log \Fdds_{t-1}(\us_t)+\sqrt{\alpha_t}(\us_{t})^{T}\nabla \log \Fdds_t(\x_{t})$. 
Then, we have 
\begin{align*}
  \zeta_{t,t-1}(\x_t,\x_{t-1})&= \log \Fdds_{t-1}(\x_{t-1})-\log \Fdds_{t-1}(\us_t)-(\x_{t-1}-\us_{t})^T\sqrt{\alpha_t}\nabla \log \Fdds_t(\x_{t})\\
  &~~~+ \log \Fdd_{t-1}(\x_{t-1})-\log \Fdds_{t-1}(\x_{t-1})\,.
\end{align*}
For a fixed $\x_t$, then we  have (the denominator is for normalization reason)
\begin{equation*}
    \Fdd_{t-1|t}(\x_{t-1}|\x_t) =\frac{\Bdds_{t-1|t}(\x_{t-1}|\x_t)\exp\bigl( \zeta_{t,t-1}(\x_t,\x_{t-1})\bigr)}{\E_{X_{t-1}\sim \Bds_{t-1|t}}\bigl[\exp\bigl( \zeta_{t,t-1}(\x_t,X_{t-1})\bigr)\bigr]}\,.
\end{equation*}
We then use the above display and Jensen's inequality to obtain  
\begin{align*}
   &\E_{X_t,X_{t-1}\sim \Fd_{t,t-1}}\biggl[\log\frac{\Fdd_{t-1|t}(X_{t-1}|X_t)}{\Bdd^s_{t-1|t}(X_{t-1}|X_t)}\biggr]\\
   &=\E_{X_t,X_{t-1}\sim \Fd_{t,t-1}}\Bigl[\zeta_{t,t-1}(X_t,X_{t-1})
   -\log \E_{X_{t-1}\sim \Bds_{t-1|t}}\bigl[\exp\bigl( \zeta_{t,t-1}(X_t,X_{t-1})\bigr)\bigr] \Bigr]\\
    &=\E_{X_t,X_{t-1}\sim \Fd_{t,t-1}}\bigl[\zeta_{t,t-1}(X_t,X_{t-1})\bigr]
   -\E_{X_t\sim \Fd_{t}}\Bigl[\log \E_{X_{t-1}\sim \Bds_{t-1|t}}\bigl[\exp\bigl( \zeta_{t,t-1}(X_t,X_{t-1})\bigr)\bigr] \Bigr]\\
   &\le \E_{X_t,X_{t-1}\sim \Fd_{t,t-1}}\bigl[\zeta_{t,t-1}(X_t,X_{t-1})\bigr]
   -\E_{X_t\sim \Fd_{t}}\Bigl[ \E_{X_{t-1}\sim \Bds_{t-1|t}}\bigl[ \zeta_{t,t-1}(X_t,X_{t-1})\bigr] \Bigr]\,.
\end{align*}
Now, let's rewrite 
\begin{align*}
  \zeta_{t,t-1}(\x_t,\x_{t-1})&= \log \Fdds_{t-1}(\x_{t-1})-\log \Fdds_{t-1}(\us_t)-(\x_{t-1}-\us_{t})^T\sqrt{\alpha_t}\nabla \log \Fdds_t(\x_{t})\\
  &~~~+ \log \Fdd_{t-1}(\x_{t-1})-\log \Fdds_{t-1}(\x_{t-1})\\
  &\eqd\zeta'_{t,t-1}(\x_t,\x_{t-1})+ \log \Fdd_{t-1}(\x_{t-1})-\log \Fdds_{t-1}(\x_{t-1})\,.
\end{align*}
We are now left with  three terms: 
1.~$\E_{X_t,X_{t-1}\sim \Fd_{t,t-1}}[\zeta'_{t,t-1}(X_t,X_{t-1})]$, 2.~$\allowbreak \E_{X_t\sim \Fd_{t},X_{t-1}\sim \Bds_{t-1|t}}[\allowbreak\zeta'_{t,t-1}(X_t,X_{t-1})]$,
and 
3.~$\E_{X_{t}\sim \Fd_{t}}[\E_{X_{t-1}\sim \Fd_{t-1|t}}[\log \Fdd_{t-1}(X_{t-1})-\log \Fdds_{t-1}(X_{t-1})]-\E_{X_{t-1}\sim \Bds_{t-1|t}}[\allowbreak\log \Fdd_{t-1}(X_{t-1})-\log \Fdds_{t-1}(X_{t-1})]]$ and we need to study  1. and 2. in details: 



\emph{Term~1:} $\allowbreak \E_{X_t\sim \Fd_{t},X_{t-1}\sim \Bds_{t-1|t}}[\zeta'_{t,t-1}(X_t,X_{t-1})]$

We  1.~use the definition of  $\zeta'_{t,t-1}(X_t,X_{t-1})$, 2.~(Second order) Taylor expand $\log \Fdds_{t-1}(X_{t-1})$ around $\us_t$, 3.~use~\cite[Lemma~7]{liang2024nonN} that implies $\E_{X_{t-1}\sim \Bds_{t-1|t}}[(X_{t-1}^i-(\us)_t^i)^p]=0~~\forall p\ge 1~odd$ (we use the notation $X_{t}^i$ to referenec to the $i$th feature of the vector $X_{t}$), 4.using the fact that $X_{t-1}^i$ is conditionally independent of $X_{t-1}^j$ for $i\ne j$ and again $\E_{X_{t-1}\sim \Bds_{t-1|t}}[(X_{t-1}^i-(\us)_t^i)^p]=0~~\forall p\ge 1~odd$, and 5.~$\E_{X_{t-1}\sim \Bds_{t-1|t}}[(X_{t-1}^i-(\us)_t^i)^2]=(1-\alpha_t)/\alpha_t$ (see \eqref{eq:mus}) 

\begin{align*}
    \E&_{X_t\sim \Fd_{t},X_{t-1}\sim \Bds_{t-1|t}}[\zeta'_{t,t-1}(X_t,X_{t-1})]\\
    &=\E_{X_t\sim \Fd_{t},X_{t-1}\sim \Bds_{t-1|t}}\bigl[\log \Fdds_{t-1}(X_{t-1})-\log \Fdds_{t-1}(\us_t)-(X_{t-1}-\us_{t})^T\sqrt{\alpha_t}\nabla \log \Fdds_t(X_{t})\bigr]\\
    &\approx \E_{X_t\sim \Fd_{t},X_{t-1}\sim \Bds_{t-1|t}}\Bigl[\nabla\log \Fdds_{t-1}(\us_t)(X_{t-1}-\us_t)-(X_{t-1}-\us_{t})^T\sqrt{\alpha_t}\nabla \log \Fdds_t(X_{t})\\
    &~~~~~~~~+\frac{1}{2}(X_{t-1}-\us_t)^T \nabla^2\log \Fdds_{t-1}(\us_t)(X_{t-1}-\us_t)\Bigr] \\
    &=\E_{X_t\sim \Fd_{t},X_{t-1}\sim \Bds_{t-1|t}}\Bigl[\frac{1}{2}(X_{t-1}-\us_t)^T \nabla^2\log \Fdds_{t-1}(\us_t)(X_{t-1}-\us_t)\Bigr] \\
    &=\frac{1}{2}\sum_{i=1}^{\Dim}  \E_{X_t\sim \Fd_{t}} \bigl[\bigl(\nabla^2\log \Fdds_{t-1}(\us_t)\bigr)_{ii}\E_{X_{t-1}\sim \Bds_{t-1|t}}\bigl(X_{t-1}^i-(\us_t)^i\bigr)^2\bigr]\\
    &=\frac{(1-\alpha_t)}{2\alpha_t}\sum_{i=1}^{\Dim}  \E_{X_t\sim \Fd_{t}} \bigl[\bigl(\nabla^2\log \Fdds_{t-1}(\us_t)\bigr)_{ii}\bigr]\,.
\end{align*}
Note that here (as well as for the Term~2), for keeping the proofs simple and tractable, we use a second-order Taylor expansion and employ the notation $\approx$.  However, higher-order expansions can also be applied without affecting the dominant rates. As we extend to higher-order Taylor expansions, the dominant factor remains  $O((1-\alpha_t)/\alpha_t)^2$, so we omit those terms for simplicity.  

\emph{Term~2:} $\E_{X_t,X_{t-1}\sim \Fd_{t,t-1}}[\zeta'_{t,t-1}(X_t,X_{t-1})]$

Following the same approach as in previous step and some further linear algebra we obtain 

\begin{align*}
    &\E_{X_t,X_{t-1}\sim \Fd_{t,t-1}}\bigl[\zeta'_{t,t-1}(X_t,X_{t-1})\bigr]\\
    &~~~~~=\E_{X_t,X_{t-1}\sim \Fd_{t,t-1}}
    \bigl[\log \Fdds_{t-1}(X_{t-1})-\log \Fdds_{t-1}(\us_t)-(X_{t-1}-\us_{t})^T\sqrt{\alpha_t}\nabla \log \Fdds_t(X_{t})\bigr]\\
    &~~~~~\approx\E_{X_t,X_{t-1}\sim \Fd_{t,t-1}}
    \Bigl[\nabla\log \Fdds_{t-1}(\us_t)(X_{t-1}-\us_t)-(X_{t-1}-\us_{t})^T\sqrt{\alpha_t}\nabla \log \Fdds_t(X_{t})\\
    &~~~~~~~~+\frac{1}{2}(X_{t-1}-\us_t)^T \nabla^2\log \Fdds_{t-1}(\us_t)(X_{t-1}-\us_t)\Bigr] \\
    &~~~~~= (1-\sqrt{\alpha_t})\E_{X_t\sim \Fd_{t}}
    \bigl[\nabla\log \Fdds_{t-1}(\us_t)\E_{X_{t-1}\sim \Fd_{t-1|t}}[(X_{t-1}-\us_t)]\bigr]\\
    &~~~~~~~~+\frac{1}{2}\E_{X_t,X_{t-1}\sim \Fd_{t,t-1}}
    \bigl[(X_{t-1}-\us_t)^T \nabla^2\log \Fdds_{t-1}(\us_t)(X_{t-1}-\us_t)\bigr]\,.
\end{align*}

Employing~\cite[Lemma~8; first claim]{liang2024nonN},  we have $\E_{X_{t-1}\sim \Fd_{t-1|t}}[X_{t-1}]=\ut$.
That implies
\begin{align*}
    \E_{X_{t-1}\sim \Fd_{t-1|t}}[(X_{t-1}-\us_t)]
    &=\E_{X_{t-1}\sim \Fd_{t-1|t}}[X_{t-1}]-\us_t\\
    &=\ut-\frac{1}{\sqrt{\alpha_t}}\bigl(\x_t+(1-\alpha_t)\nabla\log \Fdds_t(\x_t)\bigr)\\
    &=\frac{(1-\alpha_t)}{\sqrt{\alpha_t}}\bigl(\nabla\log \Fdd_t(\x_t)-\nabla\log \Fdds_t(\x_t)\bigr)\,.
\end{align*}


Collecting  the pieces above together with Cauchy–Schwarz inequality we obtain  
\begin{align*}
    &\E_{X_t,X_{t-1}\sim \Fd_{t,t-1}}\bigl[\zeta'_{t,t-1}(X_t,X_{t-1})\bigr]\\
    &~~~~~\le \frac{(1-\alpha_t)}{\sqrt{\alpha_t}}(1-\sqrt{\alpha_t})\sqrt{\E_{X_t\sim \Fd_{t}}\norm{
    \nabla\log \Fdds_{t-1}(\us_t)}^2\E_{X_t\sim \Fd_{t}}\norm{\nabla\log \Fdd_t(X_t)-\nabla\log \Fdds_t(X_t)}^2}\\
    &~~~~~~~~+\frac{1}{2}\E_{X_t,X_{t-1}\sim \Fd_{t,t-1}}
    \Bigl[(X_{t-1}-\us_t)^T \nabla^2\log \Fdds_{t-1}(\us_t)(X_{t-1}-\us_t)\Bigr]\,.
\end{align*}
Now let treat the second term in the inequality above by 1.~adding a zero-valued term, 2.~expanding the product, 3.~using~\cite[Lemma~8; second claim]{liang2024nonN} and the fact that terms two and three goes to zero and some rewriting 

   \begin{align}\label{eq:varXmus}
    \E_{X_{t}\sim \Fd_{t}}&\bigl[\E_{X_{t-1}\sim \Fd_{t-1|t}}[(X_{t-1}-\ut+\ut-\us_t)^T(X_{t-1}-\ut+\ut-\us_t)]\bigr] \notag\\
    &=\E_{X_{t}\sim \Fd_{t}}\bigl[\E_{X_{t-1}\sim \Fd_{t-1|t}}[(X_{t-1}-\ut)^T(X_{t-1}-\ut)]\bigr] \notag\\
    &+\E_{X_{t}\sim \Fd_{t}}\bigl[\E_{X_{t-1}\sim \Fd_{t-1|t}}[(X_{t-1}-\ut)^T(\ut-\us_t)]\bigr] \notag\\
    &+\E_{X_{t}\sim \Fd_{t}}\bigl[\E_{X_{t-1}\sim \Fd_{t-1|t}}[(\ut-\us_t)^T(X_{t-1}-\ut)]\bigr] \notag\\
    &+\E_{X_{t}\sim \Fd_{t}}\bigl[\E_{X_{t-1}\sim \Fd_{t-1|t}}[(\ut-\us_t)^T(\ut-\us_t)]\bigr] \notag\\
    &=\E_{X_{t}\sim \Fd_{t}}\Bigl [\frac{1-\alpha_t}{\alpha_t}\Identity+\frac{(1-\alpha_t)^2}{\alpha_t}\nabla^2 \log \Fdd_t(X_t)\Bigr]\notag\\
    &+\E_{X_{t}\sim \Fd_{t}}\biggl[\frac{(1-\alpha_t)^2}{\alpha_t}\bigl(\nabla\log\Fdds_t(X_t)-\nabla\log\Fdd_t(X_t)\bigr)^T\bigl(\nabla\log \Fdds_t(X_t)-\nabla\log \Fdd_t(X_t)\bigr)\biggr]\,.
   \end{align}
   Above results state that our term involving $(1-\alpha_t)\Identity/\alpha_t$ can be canceled out by the terms  from  Step~1. We  then need to study the remaining terms that all involve the nice factor $(1-\alpha_t)^2$. 

So, collecting all the pieces of the proof, we  1.~use the results from Term~1. and Term~2.,  2.~implying some linear algebra to expand the product, 3.~use~\eqref{eq:varXmus} and cancel out terms involving the multiple $(1-\alpha_t)$ (for simplicity we have ignored the last term in~\eqref{eq:varXmus} since it has a minor affect on our final rates), 4.~using our Assumption~\ref{Assum:approxs}, 
$\E_{X_t\sim \Fd_{t}}\norm{
    \nabla\log \Fdds_{t-1}(\us_t)}^2 \le s^2B^2 $, 5.~once again use the Assumption~\ref{Assum:approxs} that implies  $s$ sparsity between entries of $
    \nabla\log \Fdds_{t-1}(\us_t)$, that also implies sparsity for the second order derivative (it causes that just a fraction of entries get involved in those sums) (also note that the last term is appeared regarding the term that we neglected in~\eqref{eq:varXmus}), and  6.~use our Assumption over the step sizes~\eqref{eq:stepsize} to obtain 
\allowdisplaybreaks
\begingroup
     \begin{align*}
       \sum_{t=1}^T\E&_{X_t,X_{t-1}\sim \Fd_{t,t-1}}\biggl[\log\frac{\Fdd_{t-1|t}(X_{t-1}|X_t)}{\Bdd^s_{t-1|t}(X_{t-1}|X_t)}\biggr]\\
       &\le \sum_{t=1}^T\biggl(\frac{(1-\alpha_t)}{\sqrt{\alpha_t}}(1-\sqrt{\alpha_t})\sqrt{\E_{X_t}\norm{
    \nabla\log \Fdds_{t-1}(\us_t)}^2\E_{X_t}\norm{\nabla\log \Fdd_t(X_t)-\nabla\log \Fdds_t(X_t)}^2}\\
    &~~~~~~~~+\frac{1}{2}\E_{X_t,X_{t-1}\sim \Fd_{t,t-1}}
    \Bigl[(X_{t-1}-\us_t)^T \nabla^2\log \Fdds_{t-1}(\us_t)(X_{t-1}-\us_t)\Bigr]\\
       &~~~~~~~~-\frac{1}{2}\sum_{i=1}^{\Dim}  \E_{X_t\sim \Fd_{t}} \bigl(\nabla^2\log \Fdds_{t-1}(\us_t)\bigr)_{ii}\Bigl(\frac{1-\alpha_t}{\alpha_t}\Bigr)\biggr) \\
       &~~~~+\sum_{t=1}^{T}\bigl(\E_{X_{t}\sim \Fd_{t}}[\E_{X_{t-1}\sim \Fd_{t-1|t}}[\log \Fdd_{t-1}(X_{t-1})-\log \Fdds_{t-1}(X_{t-1})]\\
    &~~~~-\E_{X_{t-1}\sim \Bds_{t-1|t}}[\log \Fdd_{t-1}(X_{t-1})-\log \Fdds_{t-1}(X_{t-1})]]\bigr)\\
       &\le \sum_{t=1}^T \biggl(\frac{(1-\alpha_t)}{\sqrt{\alpha_t}}(1-\sqrt{\alpha_t})\sqrt{\E_{X_t}\norm{
    \nabla\log \Fdds_{t-1}(\us_t)}^2\E_{X_t}\norm{\nabla\log \Fdd_t(X_t)-\nabla\log \Fdds_t(X_t)}^2}\\
    &~~~~~~~~+\frac{1}{2}\E_{X_t,X_{t-1}\sim \Fd_{t,t-1}}
    \biggl[\sum_{i=1}^{\Dim}(X_{t-1}-\us_t)_i^2 \Bigl(\nabla^2\log \Fdds_{t-1}(\us_t)\Bigr)_{ii}\\   &~~~~~~~~+\sum_{i=1}^{\Dim}\sum_{j=1,j\ne i}^{\Dim}(X_{t-1}-\us_t)_i(X_{t-1}-\us_t)_j \Bigl(\nabla^2\log \Fdds_{t-1}(\us_t)\Bigr)_{ij}\Bigr]\\
       &~~~~~~~~-\frac{1}{2}\sum_{i=1}^{\Dim}  \E_{X_t\sim \Fd_{t}} \bigl(\nabla^2\log \Fdds_{t-1}(\us_t)\bigr)_{ii}\Bigl(\frac{1-\alpha_t}{\alpha_t}\Bigr)\biggr)  \\
 &~~~~+\sum_{t=1}^{T}\bigl(\E_{X_{t}\sim \Fd_{t}}[\E_{X_{t-1}\sim \Fd_{t-1|t}}[\log \Fdd_{t-1}(X_{t-1})-\log \Fdds_{t-1}(X_{t-1})]\\
    &~~~~-\E_{X_{t-1}\sim \Bds_{t-1|t}}[\log \Fdd_{t-1}(X_{t-1})-\log \Fdds_{t-1}(X_{t-1})]]\bigr)\\
       &\le \sum_{t=1}^T \biggl(\frac{(1-\alpha_t)}{\sqrt{\alpha_t}}(1-\sqrt{\alpha_t})\sqrt{\E_{X_t}\norm{
    \nabla\log \Fdds_{t-1}(\us_t)}^2\E_{X_t}\norm{\nabla\log \Fdd_t(X_t)-\nabla\log \Fdds_t(X_t)}^2}\\
    &~~~~~~~~+\frac{1}{2}\E_{X_t,X_{t-1}\sim \Fd_{t,t-1}}
    \biggl[\sum_{i=1}^{\Dim}\frac{(1-\alpha_t)^2}{\alpha_t}\bigl(\nabla^2 \log \Fdd_t(X_t)\bigr)_{ii} \Bigl(\nabla^2\log \Fdds_{t-1}(\us_t)\Bigr)_{ii}\\   &~~~~~~~~+\sum_{i=1}^{\Dim}\sum_{j=1,j\ne i}^{\Dim}\frac{(1-\alpha_t)^2}{\alpha_t}\bigl(\nabla^2 \log \Fdd_t(X_t)\bigr)_{ij} \Bigl(\nabla^2\log \Fdds_{t-1}(\us_t)\Bigr)_{ij}\Bigr]\biggr)\\
  &~~~~+\sum_{t=1}^{T}\bigl(\E_{X_{t}\sim \Fd_{t}}[\E_{X_{t-1}\sim \Fd_{t-1|t}}[\log \Fdd_{t-1}(X_{t-1})-\log \Fdds_{t-1}(X_{t-1})]\\
    &~~~~-\E_{X_{t-1}\sim \Bds_{t-1|t}}[\log \Fdd_{t-1}(X_{t-1})-\log \Fdds_{t-1}(X_{t-1})]]\bigr)\\
    &\le \sum_{t=1}^T \biggl(\frac{(1-\alpha_t)}{\sqrt{\alpha_t}}(1-\sqrt{\alpha_t})\sqrt{s^2B^2\E_{X_t\sim \Fd_{t}}\norm{\nabla\log \Fdd_t(X_t)-\nabla\log \Fdds_t(X_t)}^2}\\
    &~~~~~~~~+\frac{1}{2}\E_{X_t,X_{t-1}\sim \Fd_{t,t-1}}
    \biggl[\sum_{i=1}^{\Dim}\frac{(1-\alpha_t)^2}{\alpha_t}\bigl(\nabla^2 \log \Fdd_t(X_t)\bigr)_{ii} \Bigl(\nabla^2\log \Fdds_{t-1}(\us_t)\Bigr)_{ii}\\   &~~~~~~~~+\sum_{i=1}^{\Dim}\sum_{j=1,j\ne i}^{\Dim}\frac{(1-\alpha_t)^2}{\alpha_t}\bigl(\nabla^2 \log \Fdd_t(X_t)\bigr)_{ij} \Bigl(\nabla^2\log \Fdds_{t-1}(\us_t)\Bigr)_{ij}\Bigr]\biggr)\\
  &~~~~+\sum_{t=1}^{T}\bigl(\E_{X_{t}\sim \Fd_{t}}[\E_{X_{t-1}\sim \Fd_{t-1|t}}[\log \Fdd_{t-1}(X_{t-1})-\log \Fdds_{t-1}(X_{t-1})]\\
    &~~~~-\E_{X_{t-1}\sim \Bds_{t-1|t}}[\log \Fdd_{t-1}(X_{t-1})-\log \Fdds_{t-1}(X_{t-1})]]\bigr)\\
    &\le \sum_{t=1}^{T}\biggl(\frac{(1-\alpha_t)}{\sqrt{\alpha_t}}(1-\sqrt{\alpha_t})( sB\epsilon)+\frac{(1-\alpha_t)^2}{2\alpha_t}( s^2\DerBound^2) +\frac{(1-\alpha_t)^2}{\alpha_t}( s^2\DerBound^2)\\
    &~~~~+\frac{(1-\alpha_t)^2}{\alpha_t}(s^2\DerBound^2\epsilon)\biggr)\\
    &~~~~+\sum_{t=1}^{T}\bigl(\E_{X_{t}\sim \Fd_{t}}[\E_{X_{t-1}\sim \Fd_{t-1|t}}[\log \Fdd_{t-1}(X_{t-1})-\log \Fdds_{t-1}(X_{t-1})]\\
    &~~~~-\E_{X_{t-1}\sim \Bds_{t-1|t}}[\log \Fdd_{t-1}(X_{t-1})-\log \Fdds_{t-1}(X_{t-1})]]\bigr)\\
&\le \frac{sB \epsilon}{T}+\frac{ s^2\DerBound^2}{T} +\frac{ s^2\DerBound^2}{T}+\frac{s^2\DerBound^2\epsilon}{T}\\
    &~~~~~~~~+\sum_{t=1}^{T}\bigl(\E_{X_{t}\sim \Fd_{t}}[\E_{X_{t-1}\sim \Fd_{t-1|t}}[\log \Fdd_{t-1}(X_{t-1})-\log \Fdds_{t-1}(X_{t-1})]\\
    &~~~~~~~~-\E_{X_{t-1}\sim \Bds_{t-1|t}}[\log \Fdd_{t-1}(X_{t-1})-\log \Fdds_{t-1}(X_{t-1})]]\bigr)\,,
   \end{align*}
   \endgroup
as desired.  

\end{proof}

\subsection{Proof of Lemma~\ref{lem:emp}}\label{proof:lememp}
\begin{proof}
Our proof approach is based on the tools from empirical process theory  and  our sparsity  assumptions over the network space and $\nabla\log q^s_t(\x_t)$. 

Let start with the application of symmetrization of probabilities with $\zeta_i$ for $i\in \{1,\dots,n\}$ as i.i.d. Rademacher random variables that are independent of the data~\cite[Lemma 16.1]{van2016estimation}, and employing Contrcation principle~\cite[Theorem~4.4]{ledoux2013probability} to obtain  
\allowdisplaybreaks
\begingroup
 \begin{align*}
     \Pr \biggl(\sup_{\NetP\in \paramspaceone}\Bigl|&\E_{X_t}\norm{\scaleh\scoref(X_t,t)-\nabla\log q^s_t(X_t,t)}^2
     -\frac{1}{n} \sum_{i=1}^{n}  \norm{\scaleh\scoref(\x_t^i,t)-\nabla\log q^s_t(\x_t^i)}^2\Bigr| \ge 4\Rs\sqrt{\frac{2t}{n}}\biggr)\\
     &\le 4 \Pr \biggl(\sup_{\NetP\in \paramspaceone}\Bigl|\frac{1}{n} \sum_{i=1}^{n} \zeta_i \norm{\scaleh\scoref(\x_t^i,t)-\nabla\log q^s_t(\x_t^i)}^2\Bigr| \ge \Rs\sqrt{\frac{2t}{n}}\biggr)\\
     &\le 8 \Pr \biggl(\sup_{\NetP\in \paramspaceone}\Bigl|\frac{1}{n} \sum_{i=1}^{n} 2\zeta_i \bigl( \norm{\scaleh\scoref(\x_t^i,t)}^2+\norm{\nabla\log q^s_t(\x_t^i)}^2\bigr)\Bigr| \ge \Rs\sqrt{\frac{2t}{n}}\biggr)\\
     &\le 8 \Pr \biggl(\sup_{\NetP\in \paramspaceone}\Bigl|\frac{1}{n} \sum_{i=1}^{n} 2\zeta_i \bigl( \norm{\scoref(\x_t^i,t)}^2\Bigr| \ge \frac{\Rs}{2\scaleh^2}\sqrt{\frac{2t}{n}}\eqd t'\biggr)\\
     &~~~+8 \Pr \biggl(\sup_{\NetP\in \paramspaceone}\Bigl|\frac{1}{n} \sum_{i=1}^{n} 2\zeta_i \bigl( \norm{\nabla\log q^s_t(\x_t^i)}^2\Bigr| \ge \frac{\Rs}{2}\sqrt{\frac{2t}{n}}\eqd t''\biggr)\,.
 \end{align*}
 \endgroup
 So, we need to find tail bounds for two concentration inequalities above: 
 For the first one, we use our definition that $\normoneM{\Theta}\le 1$  and $\sup_{\NetP\in \paramspaceone} \norm{\scoref(\x_t^i,t)}_1\le 1$  that also implies  $\sup_{\NetP\in \paramspaceone} \norm{\scoref(\x_t^i,t)}^2\le 1$.
 Following  a similar uniform bound as proposed in~\cite[Proof of Theorem~5]{taheri2021} for ReLU and regularized neural networks, we can obtain 
 \begin{equation*}
      \Pr \biggl(\sup_{\NetP\in \paramspaceone}\Bigl|\frac{1}{n} \sum_{i=1}^{n} \zeta_i  \norm{\scoref(\x_t^i,t)}^2\Bigr| \ge \frac{\Rs}{4\scaleh^2}\sqrt{\frac{2t}{n}}= t'\biggr) \lessapprox \frac{1}{n}
 \end{equation*}
 for $t=((\scaleh^2+(sB)^2)(2/L)^{2L}\log(n)\sqrt{L^2\log(p)\sum_{i=1}^n\norm{\x_i}^4/n}/\mathcal{R})^2/2$, where $p$ stands for total number of network parameters and $L$ number of hidden layers of ReLU network. For simplicity, we can simplify the notation and  shortly set $t=((\scaleh^2+(sB)^2)C_{\x}\sqrt{\log(pn)}/\mathcal{R})^2/2$, where $C_{\x}$ absorbs factors related to the input and all  constants  and we also used the fact that  $(2/L)^{2L}L\le 1$ for $L\ge 3$.  
 
 For the second concentration inequality, we  use Hoeffding's inequality~\cite[Theorem~2.6.3]{Vershynin2018}  for $y_i=\zeta_i \norm{\nabla\log q^s_t(\x_t^i)}^2$, that is $y_i=\zeta_i \norm{\nabla\log q^s_t(\x_t^i)}^2$ are zero-mean random variables and bounded $\norm{\nabla\log q^s_t(\x_t^i)}^2 \le s^2 \norm{\nabla\log q^s_t(\x_t^i)}_{\infty}^2 \le s^2  \DerBound^2$, where we have used Assumption~\ref{Assum:approxs} and Assumption~\ref{assum:ReDe} to conclude that $\norm{\nabla\log q^s_t(\x_t^i)}_{\infty}\approx \norm{\nabla\log q_t(\x_t^i)}_{\infty} \le  \DerBound$.
 Now, we can progress as following: 
 \begin{align*}
     \Pr \biggl(\sup_{\NetP\in \paramspaceone}\Bigl|\E_{X_t}\norm{\scaleh\scoref(X_t,t)&-\nabla\log q^s_t(X_t)}^2
     -\frac{1}{n} \sum_{i=1}^{n}  \norm{\scaleh\scoref(\x_t^i,t)-\nabla\log q^s_t(\x_t^i)}^2\Bigr| \ge 4\Rs\sqrt{\frac{2t}{n}}\biggr)\\
     &\le 8 \Pr \biggl(\sup_{\NetP\in \paramspaceone}\Bigl|\frac{1}{n} \sum_{i=1}^{n} \zeta_i \bigl( \norm{\st(\x_t^i,t)}^2\Bigr| \ge \frac{\Rs}{4\scaleh^2}\sqrt{\frac{2t}{n}}= t'\biggr)\\
     &~~~+8 \Pr \biggl(\Bigl|\frac{1}{n} \sum_{i=1}^{n} \zeta_i \bigl( \norm{\nabla\log q^s_t(\x_t^i)}^2\Bigr| \ge \frac{\Rs}{4}\sqrt{\frac{2t}{n}}= t''\biggr)\\
     &\le \frac{16}{n}+16\exp\biggl(-\frac{n  t''^2}{c's^4\DerBound^4}\biggr)\,.
 \end{align*} 
And   using our assumptions we have  
 \begin{align*}
\Rs^2&\le  \sup_{\NetP\in \paramspaceone}\frac{1}{n}\sum_{i=1}^{n}  \E_{X_t\sim Q_t}\norm{\scaleh\scoref(X_t^i,t)-\nabla\log q^s_t(X_t^i)}^4\\
& \le 4\bigl(\scaleh^4+ s^4{\DerBound}^4\bigr)\,.
 \end{align*}
 That leaves us with 
 \begin{align*}
      \Pr \biggl(\sup_{\NetP\in \paramspaceone}\Bigl|\E_{X_t\sim Q_t}\norm{\scaleh\scoref(X_t,t)&-\nabla\log q^s_t(X_t)}^2
     -\frac{1}{n} \sum_{i=1}^{n}  \norm{\scaleh\scoref(\x_t^i)-\nabla\log q^s_t(\x_t^i)}^2\Bigr| \\
     &\ge 8\sqrt{\scaleh^4+s^4 {\DerBound}^4}\sqrt{\frac{2t}{n}}\biggr)\\
     &\le \frac{16}{n}+16\exp\biggl(-\frac{n  t''^2}{c's^4\DerBound^4}\biggr)\,.
 \end{align*}
For $t=((\scaleh^2+(sB)^2)C_{\x}\sqrt{\log(pn)}/\mathcal{R})^2/2$,  we then reach 
  \begin{align*}
      \Pr \biggl(\sup_{\NetP\in \paramspaceone}\Bigl|\E_{X_t\sim Q_t}\norm{\scaleh\scoref(X_t,t)&-\nabla\log q^s_t(X_t)}^2
     -\frac{1}{n} \sum_{i=1}^{n}  \norm{\scaleh\scoref(\x_t^i,t)-\nabla\log q^s_t(\x_t^i)}^2\Bigr| \\
     &\ge  \bigl (\scaleh^2+s^2 {\DerBound}^2\bigr)C_{\x}\sqrt{\frac{\log (pn)}{n}}\ \biggr)\\
     &\lessapprox \frac{32}{n}\,.
 \end{align*}
\end{proof}

\subsection{Proof of Lemma~\ref{lem:tilfactor}}\label{proof:tilf}
\begin{proof}
The proof is based on some simple linear algebra and the definition of forward and backward processes. 

We  1.~use Bayes' rule, 2.~consider a fixed $\x_t$ ($\Fdd_t(\x_t)$ is omitted since $\x_t$ is fixed), 3.~definition of the forward process, and 4.~multiplying with a one-valued factor and some rewriting, and 4.~and some further rewriting 
\begin{align*}
  \Fdd_{t-1|t}(\x_{t-1}|\x_t) &=\frac{\Fdd_{t|t-1}(\x_t|\x_{t-1})\Fdd_{t-1}(\x_{t-1})}{\Fdd_{t}(\x_t)} \\
  & \propto \Fdd_{t|t-1}(\x_t|\x_{t-1})\Fdd_{t-1}(\x_{t-1})\\
 & \propto \Fdd_{t-1}(\x_{t-1}) \exp \biggl(-\frac{\norm{\x_t-\sqrt{\alpha_t}\x_{t-1}}^2}{2(1-\alpha_t)}\biggr)\\
 & \propto \Bdds_{t-1|t}(\x_{t-1}|\x_t)  \exp \biggl(\log \Fdd_{t-1}(\x_{t-1})+\frac{\alpha_t\norm{\x_{t-1}-\us_t}^2}{2(1-\alpha_t)}-\frac{\norm{\x_t-\sqrt{\alpha_t}\x_{t-1}}^2}{2(1-\alpha_t)}\biggr)\\
 &=  \Bdds_{t-1|t}(\x_{t-1}|\x_t)  \exp \biggl(\log \Fdd_{t-1}(\x_{t-1})+\frac{\alpha_t\norm{\x_{t-1}-\us_t}^2}{2(1-\alpha_t)}\\
 &~~~-\frac{\alpha_t\norm{\x_{t-1}-(\x_{t}/\sqrt{\alpha_t})}^2}{2(1-\alpha_t)}\biggr)
\end{align*}
 using the fact that  (see~\eqref{eq:mus})
\begin{align*}
   \Bdds_{t-1|t}(\x_{t-1}|\x_t) & \propto \exp\biggl(-\frac{\alpha_t\norm{\x_{t-1}-\us_t}^2}{2(1-\alpha_t)}\biggr)\,.
\end{align*}
We then use the fact that 
\begin{align*}
  \norm{\x_{t-1}-\us_t}^2-\norm{\x_{t-1}&-(\x_{t}/\sqrt{\alpha_t})}^2\\
  &=\norm{\x_{t-1}-(\x_{t}/\sqrt{\alpha_t})+(\x_{t}/\sqrt{\alpha_t})-\us_t}^2-\norm{\x_{t-1}-(\x_{t}/\sqrt{\alpha_t})}^2\\
  &=2\bigl(\x_{t-1}-(\x_{t}/\sqrt{\alpha_t})\bigr)^T\bigl((\x_{t}/\sqrt{\alpha_t})-\us_t\bigr)+\norm{(\x_{t}/\sqrt{\alpha_t})-\us_t}^2\,.
\end{align*}
We then use~\eqref{eq:mus} to obtain 
\begin{align*}
 \frac{2\bigl(\x_{t-1}-(\x_{t}/\sqrt{\alpha_t})\bigr)^T\bigl((\x_{t}/\sqrt{\alpha_t})-\us_t\bigr)}{(1-\alpha_t)/\alpha_t}  & =-\bigl(\x_{t-1}-(\x_{t}/\sqrt{\alpha_t})\bigr)^T \sqrt{\alpha_t} \nabla\log \Fdds_t(\x_t)\\
 &=-\sqrt{\alpha_t}\x_{t-1}^T\nabla\log \Fdds_t(\x_t)+\x_t^T\nabla\log \Fdds_t(\x_t)\,.
\end{align*}
    
Collecting all the pieces above we obtain  
for a fixed $\x_t$ 
\begin{align*}
  \Fdd_{t-1|t}(\x_{t-1}|\x_t)  \varpropto \Bdds_{t-1|t}(\x_{t-1}|\x_t)\exp\bigl( \zeta_{t,t-1}(\x_t,\x_{t-1})\bigr)
\end{align*}
with $\zeta_{t,t-1}(\x_t,\x_{t-1})= \log \Fdd_{t-1}(\x_{t-1})-\sqrt{\alpha_t}\x_{t-1}^T\nabla \log \Fdds_t(\x_t)+f(\x_t)$, where $f(\x_t)$ can be considered as an arbitrary function of $\x_t$, since $\x_t$ was fixed (let's also note that the term $\norm{(\x_{t}/\sqrt{\alpha_t})-\us_t}^2$ is omitted since it is just dependent over $\x_t$).
That completes the proof.

\end{proof}

\subsection{Proof of Lemma~\ref{lem:logdeng}}\label{proof:loggaussian}
\begin{proof}
We employ  the fact that   $\BddA$ and $\Bdd^s$ are both Gaussian with the same variance,   $\BdA_{t-1|t}=\mathcal{N}(\x_{t-1};\uth(\x_t),\sigma_t^2\Identity)$  and 
 $\Bds_{t-1|t}=\mathcal{N}(\x_{t-1};\us_t(\x_t),\sigma_t^2\Identity)$ with 
 \begin{equation*}
 \uth(\x_t)=\frac{1}{\sqrt{\alpha_t}}\bigl(\x_t+(1-\alpha_t)\scaleh\scorefh(\x_t,t)\bigr)
\end{equation*}
and 
\begin{equation*}
    \us_t(\x_t)=\frac{1}{\sqrt{\alpha_t}}\bigl(\x_t+(1-\alpha_t)\nabla_{\x_t}\log \Fdds_t(\x_t)\bigr)\,.
\end{equation*}
We use 1.the Gaussian property, 2.~rewriting, 3.~add a zero-valued term, 4.~use the property that $\E_{X_{t-1}\sim \Fd_{t-1|t}}[X_{t-1}|\x_t]=\ut(\x_t)$, 5.~definitions of $\ut$, $\us$, and $\uth$ and Cauchy–Schwarz inequality  to obtain 
    \begin{align*}
        & \E_{X_t,X_{t-1}\sim  \Fd_{t,t-1}} \biggl[\log\frac{\Bdd^s_{t-1|t}(X_{t-1}|X_t)}{\BddA_{t-1|t}(X_{t-1}|X_t)}\biggr]\\
        &=\E_{X_t,X_{t-1}\sim \Fd_{t,t-1}} \biggl[\frac{\alpha_t}{2(1-\alpha_t)}(\norm{X_{t-1}-\uth(X_t)}^2-\norm{X_{t-1}-\us(X_t)}^2)\biggr]\\
        &=\E_{X_t,X_{t-1}\sim \Fd_{t,t-1}} \biggl[\frac{\alpha_t}{(1-\alpha_t)}\bigl(X_{t-1}-\us_t(X_t)\bigr)^{T}\bigl(\us_t(X_t)-\uth(X_t)\bigr)\\
        &~~~~~+\frac{\alpha_t}{2(1-\alpha_t)}\norm{\us_t(X_t)-\uth(X_t)}^2\biggr]\\
        &=\E_{X_t,X_{t-1}\sim \Fd_{t,t-1}} \biggl[\frac{\alpha_t}{(1-\alpha_t)}\bigl(X_{t-1}-\ut(X_t)+\ut(X_t)-\us_t(X_t)\bigr)^{T}\bigl(\us_t(X_t)-\uth(X_t)\bigr)\\
        &~~~~~+\frac{\alpha_t}{2(1-\alpha_t)}\norm{\us_t(X_t)-\uth(X_t)}^2\biggr]\\
        &=\E_{X_t,X_{t-1}\sim \Fd_{t,t-1}} \biggl[\frac{\alpha_t}{(1-\alpha_t)}\bigl(\ut(X_t)-\us_t(X_t)\bigr)^{T}\bigl(\us_t(X_t)-\uth(X_t)\bigr)\\
        &~~~~~+\frac{\alpha_t}{2(1-\alpha_t)}\norm{\us_t(X_t)-\uth(X_t)}^2\biggr]\\
    &\le (1-\alpha_t)\sqrt{\E_{X_t}\norm{\nabla\log \Fdd_t(X_t)-\nabla\log \Fdds_t(X_t)}^2\E_{X_t}\norm{\nabla\log \Fdds_t(X_t)-\scaleh\scorefh(X_t,t)}^2}\\
    &~~~+\frac{1-\alpha_t}{2}  \E_{X_t}\norm{\scaleh\scorefh(X_t,t)-\nabla\log q^s_t(X_t)}^2\,.
    \end{align*}
    That implies 
 \begin{align*}
     \sum_{t=1}^T~&\E_{X_t,X_{t-1}\sim \Fd_{t,t-1}} \biggl[\log\frac{\Bdd^s_{t-1|t}(X_{t-1}|X_t)}{\BddA_{t-1|t}(X_{t-1}|X_t)}\biggr] \\
     &\le \sum_{t=1}^T (1-\alpha_t)\sqrt{\E_{X_t}\norm{\nabla\log \Fdd_t(X_t)-\nabla\log \Fdds_t(X_t)}^2\E_{X_t}\norm{\nabla\log \Fdds_t(X_t)-\scaleh\scorefh(X_t,t)}^2}\\
    &~~~+ \sum_{t=1}^T \frac{1-\alpha_t}{2}\E_{X_t}\norm{\scaleh\scorefh(X_t,t)-\nabla\log q^s_t(X_t)}^2\,,
 \end{align*} 
 as desired.

\end{proof}

%% file: Contents/References.bib
@inproceedings{li2023towards,
  title={Towards Faster Non-Asymptotic Convergence for Diffusion-Based Generative Models},
  author={Li, G.  and Wei, Y.  and Chen, Y. and Chi, Y.},
  booktitle={Proc.\@ ICLR},
  year={2024}
}

@inproceedings{chen2023probability,
  title={The probability flow ODE is provably fast},
  author={Chen, S. and Chewi, S. and Lee, H. and Li, Y. and Lu, J. and Salim, A.},
  booktitle={Proc.\@ NIPS},
  volume={36},
  year={2023}
}

@inproceedings{chen2023improved,
  title={Improved analysis of score-based generative modeling: User-friendly bounds under minimal smoothness assumptions},
  author={Chen, H. and Lee, H. and Lu, J.},
  booktitle={Proc.\@ ICML},
  pages={4735--4763},
  year={2023}
}

@article{de2022convergence,
  title={Convergence of denoising diffusion models under the manifold hypothesis},
  author={De Bortoli, V.},
  journal={arXiv:2208.05314},
  year={2022}
}

@inproceedings{lee2022convergence,
  title={Convergence for score-based generative modeling with polynomial complexity},
  author={Lee, H. and Lu, J. and Tan, Y.},
  booktitle={Proc.\@ NIPS},
  volume={35},
  pages={22870--22882},
  year={2022}
}

@inproceedings{song2020score,
  title={Score-based generative modeling through stochastic differential equations},
  author={Song, Y. and Sohl-Dickstein, J. and Kingma, D. and Kumar, A.  and Ermon, S.  and Poole, B.},
  booktitle={Proc.\@  ICLR},
  year={2021}
}

@inproceedings{ho2020denoising,
  title={Denoising diffusion probabilistic models},
  author={Ho, J. and Jain, A. and Abbeel, P.},
  booktitle={Proc.\@ NIPS},
  volume={33},
  pages={6840--6851},
  year={2020}
}

@article{block2020generative,
  title={Generative modeling with denoising auto-encoders and Langevin sampling},
  author={Block, A. and Mroueh, Y. and Rakhlin, A.},
  journal={arXiv:2002.00107},
  year={2020}
}

@book{Lederer2021HD,
  title={Fundamentals of High-Dimensional Statistics: with exercises and R labs},
  author={Lederer, J.},
  year={2022},
  publisher={Springer Texts in Statistics}
}

@article{taheri2022balancing,
  title={Balancing Statistical and Computational Precision: A General Theory and Applications to Sparse Regression},
  author={Taheri, M.  and  Lim, N. and  Lederer, J.},
  journal={IEEE Trans.\@ Inform.\@ Theory},
  volume={69},
  number={1},
  pages={316--333},
  year={2022},
  publisher={IEEE},
  doi={10.1109/TIT.2022.3203857}
}

@article{taheri2022statistical,
  title={Statistical Guarantees for Approximate Stationary Points of Simple Neural Networks},
  author={Taheri, M. and Xie, F. and Lederer, J.},
  journal={TMLR},
  year={2025},
  note    = {Available at \url{https://openreview.net/forum?id=PNUMiLbLml}}
}

@article{vincent2011connection,
  title={A connection between score matching and denoising autoencoders},
  author={Vincent, P.},
  journal={Neural Comput.},
  volume={23},
  number={7},
  pages={1661--1674},
  year={2011},
  publisher={MIT Press}
}

@inproceedings{dhariwal2021diffusion,
  title={Diffusion models beat gans on image synthesis},
  author={Dhariwal, P. and Nichol, A.},
  booktitle={Proc.\@ NIPS},
  volume={34},
  pages={8780--8794},
  year={2021}
}

@inproceedings{song2019generative,
  title={Generative modeling by estimating gradients of the data distribution},
  author={Song, Y. and Ermon, S.},
  booktitle={Proc.\@ NIPS},
  volume={32},
  year={2019}
}

@article{ramesh2022hierarchical,
  title={Hierarchical text-conditional image generation with clip latents},
  author={Ramesh, A. and Dhariwal, P. and Nichol, A. and Chu, C. and Chen, M.},
  journal={arXiv:2204.06125},
  volume={1},
  number={2},
  pages={3},
  year={2022}
}

@inproceedings{chen2022sampling,
  title={Sampling is as easy as learning the score: theory for diffusion models with minimal data assumptions},
  author={Chen, S. and Chewi, S. and Li, J. and Li, Y. and Salim, A. and Zhang, A.},
  booktitle={Proc.\@ ICLR},
  year={2023}
}

@article{wibisono2022convergence,
  title={Convergence in KL divergence of the inexact Langevin algorithm with application to score-based generative models},
  author={Wibisono, A. and Yang, K.},
  journal={arXiv:2211.01512},
  year={2022}
}

@article{lee2023convergence,
  title={Convergence of score-based generative modeling for general data distributions},
  author={Lee, H. and Lu, J. and Tan, Y.},
  journal={ALT},
  pages={946--985},
  year={2023}
}

@inproceedings{song2021maximum,
  title={Maximum likelihood training of score-based diffusion models},
  author={Song, Y. and Durkan, C. and Murray, I. and Ermon, S.},
  booktitle={Proc.\@ NIPS},
  volume={34},
  pages={1415--1428},
  year={2021}
}

@book{Vershynin2018,
  title={High-dimensional probability: an introduction with applications in data science},
  author={Vershynin, R.},
  year={2018},
  publisher={Cambridge Univ.\@ Press}
}

@book{Sara2000,
  title={Empirical processes in M-estimation},
  author={{van de Geer}, S.},
  year={2000},
  publisher={Cambridge Univ.\@ Press}
}

@inproceedings{chen2023restoration,
  title={Restoration-degradation beyond linear diffusions: A non-asymptotic analysis for ddim-type samplers},
  author={Chen, S. and Daras, G. and Dimakis, A.},
  booktitle={Proc.\@ ICML},
  pages={4462--4484},
  year={2023},
  organization={PMLR}
}

@article{taheri2021,
  title={Statistical Guarantees for Regularized Neural Networks},
  author={Taheri,M. and Xie, F. and Lederer, J.},
 journal={Neural Networks},
  volume={142},
  pages={148--161},
  year={2021},
  publisher={Elsevier}
}

@article{lederer2023extremes,
  title={Extremes in high dimensions: Methods and scalable algorithms},
  author={Lederer, J. and Oesting, M.},
  journal={arXiv:2303.04258},
  year={2023}
}

@inproceedings{Golowich17,
  title={Size-independent sample complexity of neural networks},
  author={Golowich, N. and Rakhlin, A. and Shamir, O.},
  booktitle={Proc.\@ COLT},
pages={297--299},
  year={2018}
}

@inproceedings{song2020sliced,
  title={Sliced score matching: A scalable approach to density and score estimation},
  author={Song, Y. and Garg, S. and Shi, J. and Ermon, S.},
  booktitle={Uncertainty artif.\@ intell.},
  pages={574--584},
  year={2020},
  organization={PMLR}
}

@book{Hastie2015,
  title={Statistical learning with sparsity: The lasso and generalizations},
  author={Hastie, T. and Tibshirani, R. and Wainwright, M.},
  year={2015},
  publisher={CRC press}
}

@article{Hieber2017,
 title={Nonparametric regression using deep neural networks with ReLU activation function},
  author={Schmidt-Hieber, J.},
  journal={Ann.\@ Statist.},
  volume={48},
  number={4},
  pages={1875--1897},
  year={2020},
  publisher={Institute of Mathematical Statistics}
}

@article{Hinton2012,
  title={Deep neural networks for acoustic modeling in speech recognition: The shared views of four research groups
},
  author={Hinton, G. and Deng, L. and Yu, D. and Dahl, G. and Mohamed, A. and Jaitly, N. and Senior, A. and Vanhoucke, V. and Nguyen, P. and Sainath, T. and Kingsbury, B.},
  journal={IEEE Signal Process.\@ Mag.\@},
  volume={29},
  number ={6},
  pages={82--97},
  year={2012}
}

@inproceedings{Molchanov2017,
  title={Variational dropout sparsifies deep neural networks},
  author={Molchanov, D. and Ashukha, A. and Vetrov, D.},
  booktitle={Proc.\@ ICML},
  pages={2498--2507},
  year={2017}
}

@inproceedings{Alvarez16,
  title={Learning the number of neurons in deep networks},
  author={Alvarez, J. and Salzmann, M.},
  booktitle={Proc.\@ NIPS},
  pages={2270--2278},
  year={2016}
}

@article{Feng17,
  title={Sparse-input neural networks for high-dimensional nonparametric regression and classification},
  author={Feng, J. and Simon, N.},
  journal={arXiv:1711.07592},
  year={2017}
}

@article{hyvarinen2005estimation,
  title={Estimation of non-normalized statistical models by score matching.},
  author={Hyv{\"a}rinen, A. and Dayan, P.},
  journal={J.\@ Mach.\@ Learn.\@ Res.},
  volume={6},
  number={4},
  year={2005}
}

@article{labach2019survey,
  title={Survey of dropout methods for deep neural networks},
  author={Labach, A. and Salehinejad, H. and Valaee, S.},
  journal={arXiv:1904.13310},
  year={2019}
}

@article{gomez2019learning,
  title={Learning sparse networks using targeted dropout},
  author={Gomez, A. and Zhang, I. and Kamalakara, S. and Madaan, D. and Swersky, K. and Gal, Y. and Hinton, G.},
  journal={arXiv:1905.13678},
  year={2019}
}

@inproceedings{Neyshabur2015,
  title={Norm-based capacity control in neural networks
},
  author={Neyshabur, B. and Tomioka, R. and Srebro, N.},
  booktitle={Proc.\@ COLT},
  volume={},
  number ={},
  pages={1376--1401},
  year={2015}
}

@article{hyvarinen2007some,
  title={Some extensions of score matching},
  author={Hyv{\"a}rinen, A.},
  journal={Comput.\@ Statist.\@ Data Anal.},
  volume={51},
  number={5},
  pages={2499--2512},
  year={2007},
  publisher={Elsevier}
}

@article{lin2016estimation,
  title={Estimation of high-dimensional graphical models using regularized score matching},
  author={Lin, L. and Drton, M. and Shojaie, A.},
  journal={Electron.\@ J.\@ Stat.},
  volume={10},
  number={1},
  pages={806},
  year={2016},
  publisher={NIH Public Access}
}

@article{tibshirani1996regression,
  title={Regression shrinkage and selection via the lasso},
  author={Tibshirani, R.},
  journal={J.\@ R. Stat.\@ Soc.\@ Ser.\@ B Stat.\@ Methodol.},
  volume={58},
  number={1},
  pages={267--288},
  year={1996},
  publisher={Oxford University Press}
}

@article{mohades2023reducing,
  title={Reducing Computational and Statistical Complexity in Machine Learning Through Cardinality Sparsity},
  author={Mohades, A. and Lederer, J.},
  journal={arXiv:2302.08235},
  year={2023}
}

@book{Eldar:2012,
	Editor = {Eldar, Y. and Kutyniok, G.},
	Publisher = {Cambridge Univ.\@ Press},
	Title = {Compressed Sensing: Theory and Applications},
	Year = {2012}}

@inproceedings{de2021diffusion,
  title={Diffusion Schr{\"o}dinger bridge with applications to score-based generative modeling},
  author={De Bortoli, V. and Thornton, J. and Heng, J. and Doucet, A.},
  booktitle={Proc.\@ NIPS},
  volume={34},
  pages={17695--17709},
  year={2021}
}

@inproceedings{liang2024non,
  title={Non-asymptotic Convergence of Discrete-time Diffusion Models: New Approach and Improved Rate},
  author={Liang, Y. and Ju, P. and Liang, Y. and Shroff, N.},
  booktitle={Proc.\@ ICLR},
  year={2024}
}

@inproceedings{ho2022video,
  title={Video diffusion models},
  author={Ho, J. and Salimans, T. and Gritsenko, A. and Chan, W. and Norouzi, M. and Fleet, D.},
  booktitle={Proc.\@ NIPS},
  volume={35},
  pages={8633--8646},
  year={2022}
}

@article{xu2022geodiff,
  title={Geodiff: A geometric diffusion model for molecular conformation generation},
  author={Xu, M. and Yu, L. and Song, Y. and Shi, C. and Ermon, S. and Tang, J.},
  journal={arXiv:2203.02923},
  year={2022}
}

@book{ledoux2013probability,
  title={Probability in Banach Spaces: isoperimetry and processes},
  author={Ledoux, M. and Talagrand, M.},
  year={2013},
  publisher={Springer Science \& Business Media}
}

@article{wibisono2024optimal,
  title={Optimal score estimation via empirical bayes smoothing},
  author={Wibisono, A. and Wu, Y. and Yang, K.},
  journal={arXiv:2402.07747},
  year={2024}
}

@article{zhang2024minimax,
  title={Minimax optimality of score-based diffusion models: Beyond the density lower bound assumptions},
  author={Zhang, K. and Yin, C. and Liang, F. and Liu, J.},
  journal={arXiv:2402.15602},
  year={2024}
}

@article{dou2024optimal,
  title={From optimal score matching to optimal sampling},
  author={Dou, Z. and Kotekal, S. and Xu, Z. and Zhou, H.},
  journal={arXiv:2409.07032},
  year={2024}
}

@article{liang2024nonN,
  title={Broadening Target Distributions for Accelerated Diffusion Models via a Novel Analysis Approach},
  author={Liang, Y. and Ju, P. and Liang, Y. and Shroff, N.},
  journal={arXiv:2402.13901},
  year={2024}
}

@article{li2024accelerating,
  title={Accelerating convergence of score-based diffusion models, provably},
  author={Li, G. and Huang, Y. and Efimov, T. and Wei, Y. and Chi, Y. and Chen, Y.},
  journal={arXiv:2403.03852},
  year={2024}
}

@article{huang2024convergence,
  title={Convergence analysis of probability flow ODE for score-based generative models},
  author={Huang, D. and Huang, J. and Lin, Z.},
  journal={arXiv:2404.09730},
  year={2024}
}

@book{van2016estimation,
  title={Estimation and testing under sparsity},
  author={van de Geer, S.},
  year={2016},
  publisher={Springer}
}

@inproceedings{LeCun1998,
  title={Gradient-based learning applied to document recognition},
  author={LeCun, Y. and Bottou, L. and Bengio, Y. and Haffner, P.},
  booktitle={Proc.\@ IEEE},
  pages={2278--2324},
  year={1998}
}

@article{Xiao2017,
  title={Fashion-{MNIST}: a novel image dataset for benchmarking machine learning algorithms},
  author={Xiao, H. and Rasul, K. and Vollgraf, R.},
  journal={arXiv:1708.07747},
  year={2017}
}

@article{benton2023linear,
  title={Linear convergence bounds for diffusion models via stochastic localization},
  author={Benton, J. and De Bortoli, V. and Doucet, A. and Deligiannidis, G.},
  journal={Proc.\@ ICLR},
  year={2024}
}

@inproceedings{chen2023score,
  title={Score approximation, estimation and distribution recovery of diffusion models on low-dimensional data},
  author={Chen, M. and Huang, K. and Zhao, T. and Wang, M.},
  booktitle={Proc.\@ ICML},
  pages={4672--4712},
  year={2023},
  organization={PMLR}
}

@article{HEBIRI2025106195,
	title = {Layer sparsity in neural networks},
	journal = {J.\@ Statist.\@ Plann.\@ Inference},
	volume = {234},
	pages = {106195},
	year = {2025},
	issn = {0378-3758},
	author = {Hebiri, M. and  Lederer, J.  and Taheri, M.},
    doi={10.1016/j.jspi.2024.106195}
}

@article{golestaneh2024many,
	title={How many samples are needed to train a deep neural network?},
	author={Golestaneh, P.  and Taheri, M.  and Lederer, J.},
	journal={Proc.\@ ICLR},
	year={2025}
}

@inproceedings{karras2024analyzing,
  title={Analyzing and improving the training dynamics of diffusion models},
  author={Karras, T. and Aittala, M. and Lehtinen, J. and Hellsten, J. and Aila, T. and Laine, S.},
  booktitle={CVPR},
  pages={24174--24184},
  year={2024}
}

@inproceedings{zhu2023sample,
  title={Sample complexity bounds for score-matching: Causal discovery and generative modeling},
  author={Zhu, Z. and Locatello, F. and Cevher, V.},
  booktitle={Proc.\@ NIPS},
  volume={36},
  pages={3325--3337},
  year={2023}
}

@article{baptista2025memorization,
  title={Memorization and Regularization in Generative Diffusion Models},
  author={Baptista, R. and Dasgupta, A. and Kovachki, N. and Oberai, A. and Stuart, A.},
  journal={arXiv preprint arXiv:2501.15785},
  year={2025}
}

@article{gabriel2025kernel,
  title={Kernel-Smoothed Scores for Denoising Diffusion: A Bias-Variance Study},
  author={Gabriel, F. and Ged, F.  and Veiga, M.  and Schertzer, E.},
  journal={arXiv preprint arXiv:2505.22841},
  year={2025}
}

@inproceedings{gupta2024improved,
  title={Improved sample complexity bounds for diffusion model training},
  author={Gupta, S.  and Parulekar, A.  and Price, E.  and Xun, Z.},
  booktitle={Proc.\@ NIPS},
  volume={37},
  pages={40976--41012},
  year={2024}
}

@article{ren2025unified,
  title={A unified approach to analysis and design of denoising markov models},
  author={Ren, Y. and Rotskoff, G. and Ying, L.},
  journal={ arXiv:2504.01938},
  year={2025}
}

@article{Gitte2025,
	title={Non-asymptotic convergence of probability flow ODEs under
	weak log-concavity
	},
	author={Kremling, G. and Iafrate, F. and Taheri, M. and  Lederer, J.},
	journal={arXiv:2510.17608},
	volume={},
	pages={},
	year={2025},
    note    = {Available at \url{https://arxiv.org/pdf/2510.17608?}}
}
